\documentclass[11pt]{article}

\usepackage[final]{acl}

\usepackage{times}
\usepackage{latexsym}

\usepackage[T1]{fontenc}
\usepackage[utf8]{inputenc}

\usepackage{microtype}

\usepackage{inconsolata}

\usepackage{graphicx}

\usepackage{hyperref}
\usepackage{url}

\usepackage{booktabs}   
\usepackage{caption} % professional-quality tables
\usepackage{amsfonts}       % blackboard math symbols
\usepackage{amsmath,amssymb}
\usepackage{nicefrac}       % compact symbols for 1/2, etc.
\usepackage{microtype}      % microtypography
\usepackage{xcolor}         % colors
\usepackage{amsmath,calc}
\usepackage{enumitem}
\usepackage{graphicx} 
\usepackage{comment}
\usepackage[most]{tcolorbox}
\usepackage{makecell}
\usepackage{lipsum}
\usepackage{multirow}
\usepackage{subcaption}
\usepackage{colortbl}
\usepackage{arydshln}
\usepackage{wrapfig}
\usepackage{floatflt}

\usepackage{xspace}

\makeatletter
\DeclareRobustCommand\onedot{\futurelet\@let@token\@onedot}
\def\@onedot{\ifx\@let@token.\else.\null\fi\xspace}
\def\eg{\emph{e.g}\onedot}

\makeatother

\title{Learning Deliberately, Acting Intuitively: \\Unlocking Test-Time Reasoning in Multimodal LLMs}

\author{
 \textbf{Yahan Yu\textsuperscript{1}\thanks{This work was done when Yahan Yu was interning at NEC Corporation, Japan.}},
 \textbf{Yuyang Dong\textsuperscript{2}},
 \textbf{Masafumi Oyamada\textsuperscript{3}},
\\
 \textsuperscript{1}Kyoto University 
 \textsuperscript{2}Initial S 
 \textsuperscript{3}NEC Corporation, Japan 
\\
\texttt{yahan@nlp.ist.i.kyoto-u.ac.jp}, \texttt{dongyuyang@initial-s.com}, \texttt{stillpedant@gmail.com}}

\begin{document}
\maketitle

\begin{abstract}
Reasoning is essential for large language models (LLMs), especially in complex tasks such as mathematical problem solving. However, multimodal reasoning still faces challenges in modality alignment and training scalability, as many existing methods rely on additional annotations or complex rule-based rewards. To address these issues, we propose the \textbf{D}eliberate-to-\textbf{I}ntuitive reasoning framework (D2I), which improves the understanding and reasoning abilities of multimodal LLMs (MLLMs) without extra annotations or complex rewards. During training, D2I uses deliberate reasoning strategies supervised only by rule-based format rewards to enhance modality alignment. During inference, it shifts to intuitive reasoning by removing these explicit strategies, allowing the model to implicitly apply the acquired abilities in its responses. D2I outperforms baselines on both in-domain and out-of-domain benchmarks, highlighting the effectiveness of format rewards in fostering transferable multimodal reasoning skills and suggesting the benefit of decoupling training-time reasoning depth from test-time response flexibility.
% Reasoning is a key capability for large language models (LLMs), particularly when applied to complex tasks such as mathematical problem solving. 
% However, multimodal reasoning research still requires further exploration of modality alignment and training costs. Many of these approaches rely on additional data annotation and relevant rule-based rewards to enhance the understanding and reasoning ability, which significantly increases training costs and limits scalability. To address these challenges, we propose the \textbf{D}eliberate-to-\textbf{I}ntuitive reasoning framework (D2I) that improves the understanding and reasoning ability of multimodal LLMs (MLLMs) without extra annotations and complex rewards. Specifically, our method sets deliberate reasoning strategies to enhance modality alignment only through the rule-based format reward during training. 
% While evaluating, the reasoning style shifts to intuitive, which removes deliberate reasoning strategies during training and implicitly reflects the model's acquired abilities in the response. D2I outperforms baselines across both in-domain and out-of-domain benchmarks. Our findings highlight the role of format reward in fostering transferable reasoning skills in MLLMs, and inspire directions for decoupling training-time reasoning depth from test-time response flexibility.
\end{abstract}

\section{Introduction}

\begin{figure}[t]
\begin{center}
\centering
\includegraphics[width=\columnwidth]{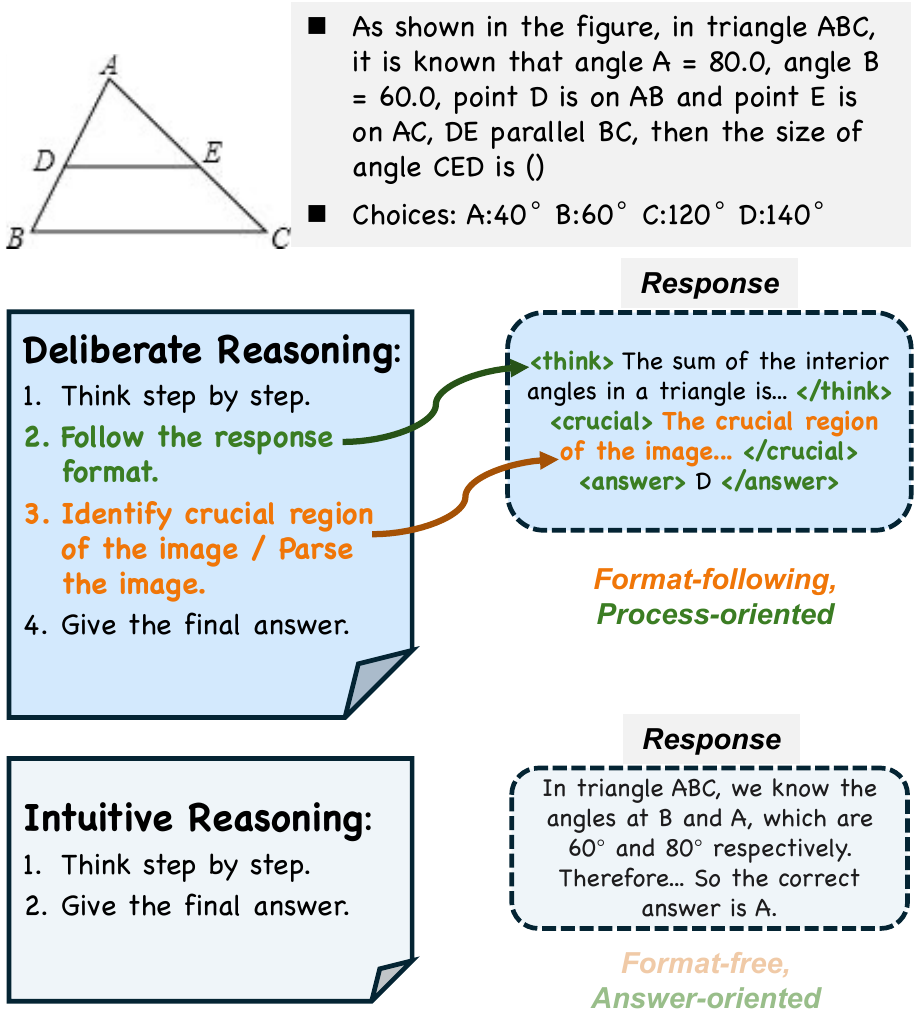}
\end{center}
\caption{The concept of the task, our designed deliberate reasoning, and intuitive reasoning.}
\label{fig:page1}
\end{figure}

% Recently, a growing body of research has distinguished between fast-thinking models \citep{cheng2025think,li2025system}, which rely on immediate pattern recognition, and slow-thinking models \citep{dou2025dsadf}, which explicitly simulate step-by-step reasoning. As the requirements of large language models (LLMs) \citep{zhang2024mm} increase for complex tasks, such as mathematical problem-solving, their ability to perform long reasoning processes and slow thinking becomes a key factor in achieving reliable performance. Recent works, such as OpenAI's o1 series \citep{jaech2024openai}, adopt the slow-thinking paradigm by scaling up inference-time Chain-of-Thought (CoT) \citep{xia2024beyond} reasoning, yielding notable improvements in tasks such as mathematics and code generation. After that, DeepSeek-R1 \citep{guo2025deepseek} demonstrates that reinforcement learning (RL) \citep{chen2025learning} can further enhance slow-thinking behavior by optimizing reasoning-specific objectives beyond standard supervised learning. 

Recent studies distinguish fast-thinking models \citep{cheng2025think,li2025system}, which rely on immediate pattern recognition, from slow-thinking models \citep{dou2025dsadf}, which conduct explicit step-by-step reasoning. As LLMs \citep{zhang2024mm} tackle more complex tasks, long-form reasoning has become essential for reliable performance. OpenAI's o1 series \citep{jaech2024openai} improves mathematics and code generation by scaling inference-time Chain-of-Thought reasoning \citep{xia2024beyond}, while DeepSeek-R1 \citep{guo2025deepseek} further enhances slow-thinking through reinforcement learning \citep{chen2025learning}.

\begin{figure*}[th!]
\begin{center}
\centering
\includegraphics[width=0.95\textwidth]{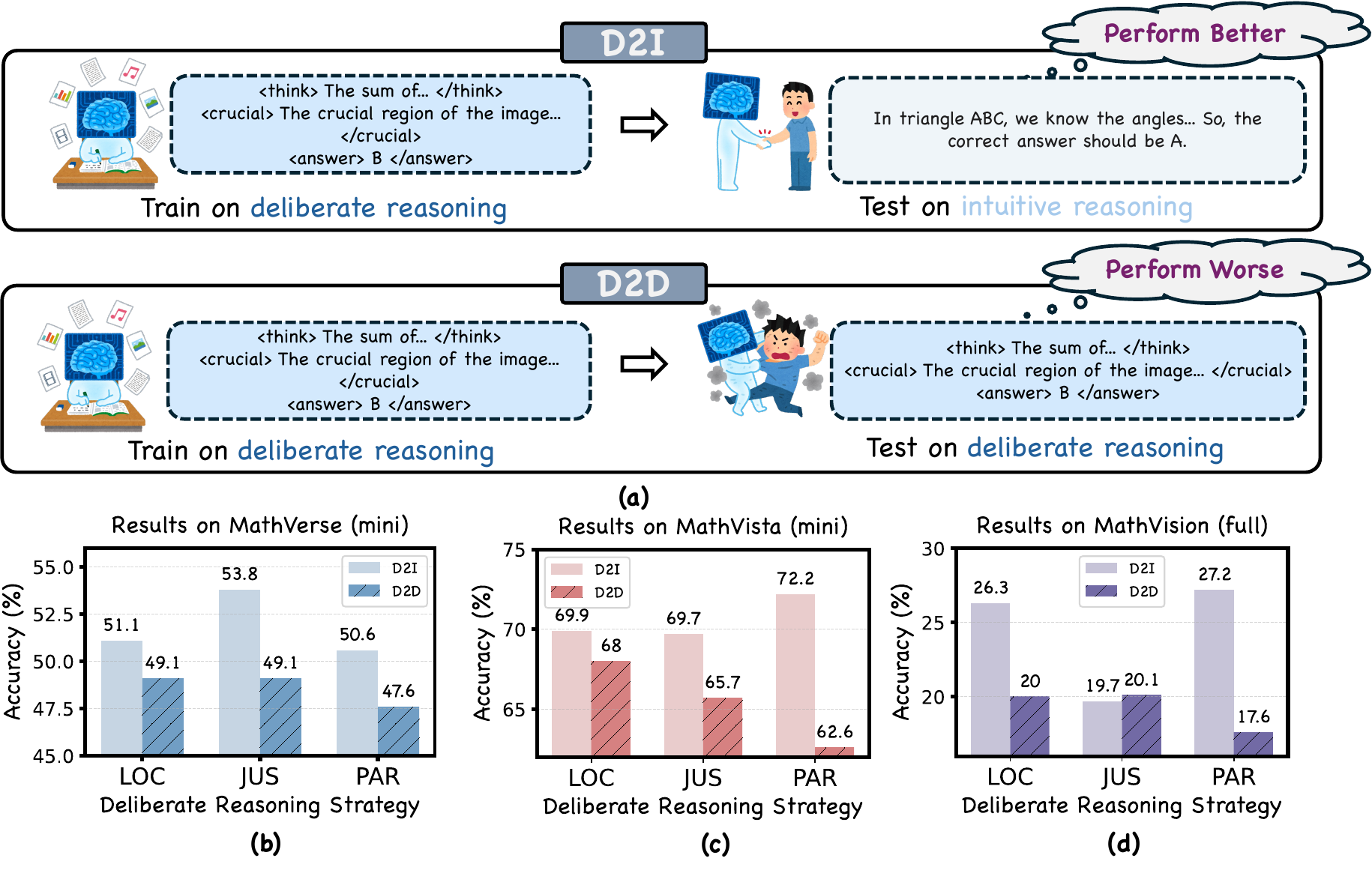}
\end{center}
\caption{\textbf{(a)} The training and testing paradigm of our proposed deliberate-to-intuitive (D2I) framework. We refer to the commonly used training-testing framework in other works as D2D. \textbf{(b)(c)(d)} The performance on mainstream benchmarks. Our D2I with different deliberate reasoning strategies consistently outperforms D2D, illustrating that the reasoning ability is unlocked during test time. }
\label{fig:teaser}
\end{figure*}

However, despite their significant advancements, these models still suffer from limited multimodal reasoning performance. Two major challenges underlie this limitation. The first is \textbf{multimodal alignment}: Accurate alignment serves as a fundamental prerequisite for reasoning in multimodal LLMs (MLLMs) \citep{zheng2024towards,DBLP:journals/corr/abs-2305-06500}. In complex tasks such as multimodal mathematics \citep{wang2024measuring}, where fine-grained visual clues must be correctly identified, any misalignment between modalities undermines the model’s ability to reason meaningfully. The second challenge concerns the \textbf{training efficiency and scalability of reasoning supervision}. Many recent approaches \citep{peng2024multimath,yang2025r1} rely on additional data annotation or complex rule-based rewards to guide models toward better reasoning behavior. While effective, these strategies significantly increase training costs and limit scalability, particularly in multimodal settings where annotation is both expensive and ambiguous. As a result, the development of lightweight and generalizable training strategies for multimodal reasoning remains an open and important direction.

% To address these challenges, our goal is to develop an R1-style visual reasoning model that excels in complex, multimodal reasoning tasks. Specifically, we aim to equip MLLMs with stronger multimodal understanding and reasoning abilities through lightweight training strategies. Inspired by the distinction between training and inference phases, we observe that while the training process should encourage the model to acquire new capabilities (\eg, understanding and analyzing images) through aggressive supervision and relatively smaller response search space, the inference process should focus on allowing a larger search space to explore more effective solutions by imposing minimal constraints (\eg, achieving the training objectives implicitly), thereby producing reliable and high-confidence responses with the acquired capabilities.

To address these challenges, our goal is to develop an R1-style visual reasoning model that improves complex multimodal reasoning through lightweight training strategies. As shown in Fig.\ref{fig:page1}, we define and distinguish between \textbf{deliberate reasoning}, which explicitly guides the model to conduct fine-grained visual understanding and reasoning, and \textbf{intuitive reasoning}, which allows flexible responses without explicit intermediate constraints. Based on this distinction, we think that training should encourage deliberate reasoning through stronger supervision and a constrained response space, enabling the model to acquire capabilities such as visual grounding and structural analysis. At inference time, however, the model should reason more intuitively, with fewer constraints and a larger response search space, so that the acquired capabilities can be applied implicitly to produce reliable responses.

To this end, we propose the \textbf{D}eliberate-\textbf{to}-\textbf{I}ntuitive \textbf{(D2I)} reasoning framework, as shown in Fig.\ref{fig:teaser}. D2I encourages deliberate reasoning during training using simple rule-based rewards and response-format constraints, while enabling intuitive reasoning at inference by removing these constraints. Beyond standard GRPO-style training \citep{guo2025deepseek}, D2I incorporates deliberate visual reasoning strategies, such as localizing key image regions and parsing visual structures, to enhance multimodal understanding without extra human annotations or content-level supervision. In contrast, we denote the commonly used setting that applies deliberate reasoning in both training and inference as D2D \citep{zhao2025boosting,zhang2023multimodal}. Our results show that such format-only supervision during training is sufficient to improve both in- and out-of-domain multimodal reasoning.

Our contributions can be summarized as follows:
\begin{itemize}
    \item We propose D2I, a reinforcement learning framework that enhances the reasoning ability of MLLMs by employing deliberate reasoning during training and intuitive reasoning during inference.

    \item To enhance the multimodal understanding ability for better reasoning, corresponding deliberate reasoning strategies are designed during training. 
    
    \item For efficient reasoning, we didn't annotate the dataset and only supervised the response format.

    \item Experiments demonstrate the effectiveness of our D2I on both in-domain and out-of-domain multimodal benchmarks.
\end{itemize}

\section{Related Work}\label{relatedwork}

%-----------------------------------------------------------------------
\subsection{MLLMs}
Multimodal large language models (MLLMs) are an extension of large language models (LLMs)~\citep{chung2022scaling,bai2023qwen,touvron2023llama2,li2025system}, enhancing their ability to process both textual and visual inputs. By combining the strong reasoning capabilities of LLMs with the rich visual features extracted by vision backbones, these models demonstrate advanced multimodal reasoning and deeper content understanding\citep{zhang2024mm,zhao2025chartedit}. Some approaches, such as LLaVA~\citep{liu2023llava}, NExT-GPT~\citep{wu2023next}, and MiniGPT-v2~\citep{chen2023minigpt}, use a linear projection layer to connect a frozen LLM with a visual encoder, enabling multimodal alignment. In contrast, other methods like InstructBLIP~\citep{DBLP:journals/corr/abs-2305-06500} and BLIP-2~\citep{DBLP:conf/icml/0008LSH23} train a dedicated Q-Former projection module to bridge the gap between different modalities.
Together, these developments showcase the rapid progress of MLLMs in tackling cross-modal reasoning and comprehension tasks.
%-----------------------------------------------------------------------

%-----------------------------------------------------------------------
\subsection{LLM Reasoning}
Reasoning in LLMs \citep{yang2025r1} refers to the ability to perform multi-step inference, often by simulating a process of deliberate thinking. This capability is particularly important for solving complex tasks such as mathematical problem solving \citep{tong2024dart}, where simple pattern matching or shallow associations are insufficient. In this context, researchers have increasingly drawn inspiration from the psychological framework of fast thinking versus slow thinking (System 1 vs. System 2), where slow thinking represents a more reflective problem-solving process \citep{zavolokina2024think}.
OpenAI’s o1 models \citep{jaech2024openai} marked a significant milestone in this direction by introducing test-time scaling, where the reasoning process is extended via longer CoT prompting. DeepSeek-R1 \citep{guo2025deepseek} adopts a reinforcement learning framework that improves the model's reasoning ability. By optimizing rule-based rewards \citep{shao2024deepseekmath}, R1 guides the model toward producing well-structured reasoning traces. Our work is inspired by these remarkable reasoning works and aims to further investigate a more appropriate and flexible training framework for MLLMs.
%-----------------------------------------------------------------------

%-----------------------------------------------------------------------
\section{Preliminary} \label{sec:3}
%-----------------------------------------------------------------------

%-----------------------------------------------------------------------
\subsection{Problem Definition} \label{sec:3.1}
%-----------------------------------------------------------------------

%-----------------------------------------------------------------------
% We focus on challenging multimodal reasoning tasks \citep{zhang2024mathverse,wang2024measuring,lu2023mathvista}, particularly those involving mathematical problem solving that requires integrating both visual and textual information. In this setting, each input instance consists of a natural language question paired with a relevant image, such as a diagram, chart, or visualized equation. The goal is to produce an output that not only includes the final answer, but also provides a text-based reasoning process that reflects the intermediate steps leading to the solution.

We focus on challenging multimodal reasoning tasks \citep{zhang2024mathverse,wang2024measuring,lu2023mathvista}, particularly mathematical problem solving over paired text-image inputs such as diagrams, charts, or visualized equations. The model is expected to generate both the final answer and a text-based reasoning process that explains the intermediate solution steps. 
Formally, given an input pair $x = (x^{\text{text}}, x^{\text{img}})$, where $x^{\text{text}}$ is the textual question and $x^{\text{img}}$ is the associated image, the model is expected to generate a response $y = (y^{\text{sol}}, y^{\text{ans}})$. Here, $y^{\text{sol}}$ denotes a coherent reasoning trace in natural language, and $y^{\text{ans}}$ is the final answer. 
%-----------------------------------------------------------------------

%-----------------------------------------------------------------------
\subsection{GRPO} \label{sec:3.2}
%-----------------------------------------------------------------------

%-----------------------------------------------------------------------
GRPO \citep{guo2025deepseek} is an efficient variant of Proximal Policy Optimization (PPO) \citep{schulman2017proximal}, designed to reduce computational overhead while maintaining competitive learning effectiveness. 
The core idea of GRPO is to sample multiple responses for the same input and evaluate their relative quality. Instead of computing absolute rewards, the method calculates the groupwise advantage by comparing each response to the average performance within the group.  
The objective function is:

\begin{equation}
\begin{aligned}
    \mathcal{J} _{GRPO} &=  \mathbb{E}[\frac{1}{N}{\textstyle \sum_{i}^{N}}min(d_iA_i, cA_i-\beta\cdot \text{KL})] \\
    c &= clip(d_i,1-\epsilon,1+\epsilon)
\end{aligned}
\end{equation}

%------------------------------------%
\begin{figure*}[t]
    \begin{center}
        \includegraphics[width=0.95\textwidth]{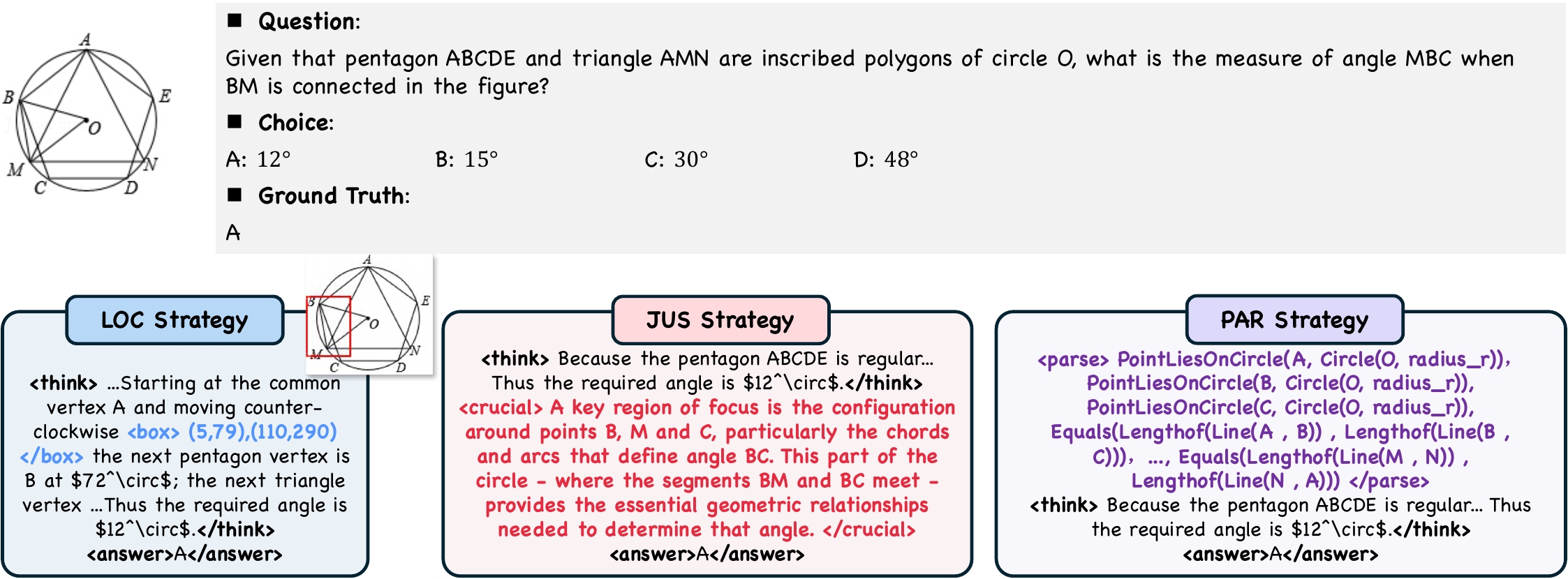}
    \end{center}
	\caption{Our proposed three deliberate reasoning strategies.}
	\label{fig:pipeline}
\end{figure*}
%------------------------------------%

Note that GRPO incorporates two types of rule-based rewards. The first is the format reward, which encourages the LLM to structure its response by enclosing the reasoning process within \texttt{<think></think>} and the final answer within \texttt{<answer></answer>}. The second is answer reward, which rewards responses where content enclosed within \texttt{<answer></answer>} matches the correct final answer. The final reward $r_i$ is computed as the average of these two components.
% GRPO supports flexible reward design, including format reward whichand answer reward. For example, in DeepSeek-R1 \citep{guo2025deepseek}, rewards are used for evaluating LLM's response, including special tags (\eg, \texttt{<think>} and \texttt{<answer>}) and correct final answers.
In our setting, reasoning process may also involve identifying or referencing specific visual elements (\eg, bounding box coordinates), requiring model to ground its reasoning in vision modality.

%-----------------------------------------------------------------------
\section{Approach} \label{sec:4}
%-----------------------------------------------------------------------

%-----------------------------------------------------------------------
\subsection{Overview}

% We propose the D2I reasoning framework to address the limitations of existing multimodal reasoning approaches. D2I is motivated by the observation that training-time reasoning can be more deliberate and structured for aggressive updating to enhance the understanding of multimodal input, while inference-time behavior should remain flexible and intuitive for stable performance. The key idea is to supervise deep reasoning behaviors during training, and allow unconstrained generation during inference. For the training-inference framework commonly adopted in most existing works, we refer to it as D2D, where both the training and inference stages are characterized by deliberate reasoning behaviors.

% Within our framework, we design three types of deliberate reasoning strategies, each encouraging the model to engage in structured reasoning grounded in visual semantics: \emph{\textbf{Region Localization (LOC) Strategy, Region Justification (JUS) Strategy and Parsing Consistency (PAR) Strategy}}. These strategies target different aspects to enforce through rule-based format reward, without requiring additional human annotations. Details are shown as follows. While implementation, we keep the accuracy reward in GRPO and only adjust the corresponding format reward.

We propose the D2I reasoning framework to improve multimodal reasoning by separating training-time and inference-time reasoning behaviors. D2I encourages deliberate and structured reasoning during training to strengthen multimodal understanding, while allowing flexible and intuitive generation during inference for stable performance. In contrast, we refer to the commonly used setting that applies deliberate reasoning in both training and inference as D2D.

Within D2I, we design three deliberate reasoning strategies: \emph{\textbf{Region Localization (LOC), Region Justification (JUS), and Parsing Consistency (PAR)}}. Each strategy promotes visually grounded structured reasoning through rule-based format rewards, without requiring additional human annotations. Following GRPO, we keep accuracy reward unchanged and only modify the format reward.

%-----------------------------------------------------------------------

%-----------------------------------------------------------------------
%-----------------------------------------------------------------------

%-----------------------------------------------------------------------
\subsection{Deliberate Reasoning Strategies} \label{sec:4.1}
%-----------------------------------------------------------------------

%-----------------------------------------------------------------------
\noindent \textbf{Region Localization (LOC) Strategy.} LOC encourages the model to identify image region most relevant to reasoning by outputting its coordinates, such as a bounding box. The format reward is triggered when valid coordinates appear within the designated tag, \eg, \texttt{<box>} $(5,79),(110,290)$ \texttt{</box>}, guiding the model to localize before reasoning. As shown in Fig.~\ref{fig:pipeline}, the expected response format is:
% The LOC strategy encourages the model to identify which part of the image is most relevant to the reasoning process by explicitly outputting the coordinates (\eg, bounding box) of that region. The corresponding format reward is triggered when a coordinate appears within the designated tag (\ie, \texttt{<box>} $(5, 79), (110,290)$ \texttt{</box>}) inside the reasoning trace. This objective guides the model to first locate before reasoning. 
% As shown in Fig.~\ref{fig:pipeline}, we expect the model to generate responses in the following format: 

\begin{tcolorbox}
\texttt{<think>} \textit{reasoning here} \texttt{<box>} \textit{coordinate here} \texttt{</box>} \textit{reasoning here} \texttt{</think>} \texttt{<answer>} \textit{final answer here} \texttt{</answer>}
\end{tcolorbox}

\noindent \textbf{Region Justification (JUS) Strategy.} JUS encourages the model to explain crucial visual clues in natural language during reasoning. The format reward is assigned when the designated tag, \texttt{<crucial>}...\texttt{</crucial>}, contains a coherent description of relevant visual elements, promoting interpretability and visual-semantic alignment. As shown in Fig.~\ref{fig:pipeline}, the expected response format is:
% The JUS strategy targets the explanation of crucial visual clues. Specifically, the model is encouraged to describe the image regions crucial for solving the question in natural language as part of its reasoning process. The format reward is assigned when the designated tag (\ie, \texttt{<crucial>}...\texttt{</crucial>}) contains a coherent textual description that refers to visual elements. This supports interpretability and encourages visual-semantic alignment. As shown in Fig.~\ref{fig:pipeline}, we expect the model to generate responses in the following format: 

\begin{tcolorbox}
\texttt{<think>} \textit{reasoning here} \texttt{</think>} \texttt{<crucial>} \textit{textual explanation here} \texttt{</crucial>} \texttt{<answer>} \textit{final answer here} \texttt{</answer>}
\end{tcolorbox}

%-----------------------------------------------------------------------

% Main results of our proposed GEOQA-8K-trained D2I method on both in-domain and out-of-domain test sets. The \textbf{bold} denotes the best performance, while the \underline{underline} denotes the second best one. In the Qwen w/ GRPO, we evaluate the model with the deliberate reasoning, while in w/ GRPO$^{\dagger}$ we evaluate with the intuitive reasoning. ${\Delta}_{base}$ represents the improvement of D2I over Qwen2.5-VL-7B, while ${\Delta}_{grpo\dagger}$ represents the improvement over Qwen2.5-VL-7B w/ GRPO$^{\dagger}$. $^{\star}$ denotes our re-implementation.

\begin{table*}[t]
\centering

	\resizebox{\linewidth}{!}{
		\begin{tabular}{lccccccccc}
			\toprule
            % \arrayrulecolor{black}
			 \multirow{2}{*}{\bf Method} & \multicolumn{1}{c}{\textbf{In-domain}} & \multicolumn{3}{c}{\textbf{Out-of-domain (Math)}} & \multicolumn{5}{c}{\textbf{Out-of-domain (General)}} \\ 
               \cmidrule(r){2-2}  \cmidrule(r){3-5} \cmidrule(r){6-10} \noalign{\smallskip} 
             % \hhline{~---------}
             % \noalign{\vskip.2pt}
			  & \multicolumn{1}{c}{\textbf{GEOQA-8K}} & \textbf{\makecell{MathVerse \\ (mini)}} & \textbf{\makecell{MathVista \\ (mini)}} & \textbf{\makecell{MATH-Vision \\ (full)}} & \textbf{\makecell{MME \\ (sum)}} & \textbf{\makecell{MMVet \\ (turbo)}} & \textbf{\makecell{MMMU \\ (val)}} & \textbf{SEED} & \textbf{POPE} \\ \hline

              \rowcolor{blue!5} \multicolumn{10}{c}{\textit{Closed-Source General Models}}\\ \hline

              GPT-4V \citep{gpt4v} & \multicolumn{1}{c}{--} & 39.4 & 58.1 & 22.7 & 1926.6 & 67.5 & 63.1 & 53.8 & -- \\
              GPT-4o \citep{hurst2024gpt} & \multicolumn{1}{c}{--} & 50.2 & 63.8 &\textbf{30.3} & -- & \underline{69.1} & \textbf{69.1} & 72.0 & 86.9 \\ \hline
              \rowcolor{blue!5} \multicolumn{10}{c}{\textit{Open-Source General Models}}\\ \hline
              Qwen2-VL-7B \citep{bai2023qwen} & \multicolumn{1}{c}{--} & 31.9 & 58.2 & 16.3 & 2326.8 & 62.0 & 54.1 & 75.1 & 88.1 \\
              InternVL2-8B \citep{chen2024far} & -- & 37.0 &58.3&18.4&2210.3&54.2&52.6 & --& 86.9 \\
              InternVL2.5-8B \citep{chen2024expanding} & \multicolumn{1}{c}{--} & 39.5 & 64.4 & 19.7 & 2344.1 & 62.8& 56.0 & -- & \textbf{90.6}  \\

              \hline
              \rowcolor{blue!5} \multicolumn{10}{c}{\textit{Reasoning Models}}\\ \hline
              LLaVA-CoT-11B \citep{xu2024llava} & \multicolumn{1}{c}{--} & 20.3  & 54.8 & -- & -- & 60.3 & -- & -- & --  \\
              
              R1-Onevision-7B \citep{yang2025r1} & \multicolumn{1}{c}{--} & 46.4 & 64.1 & -- & 2192.2 & 67.5 & -- &  66.5 & 84.9 \\
              OpenVLThinker-7B \citep{deng2025openvlthinker} & \multicolumn{1}{c}{--} & 47.9 & \underline{70.2} & 25.3 & --& --& --& --& -- \\
              % MM-Eureka-7B & \multicolumn{1}{c}{--} & 50.3 & 73.0 & -- & --& --& --& --& -- \\
              \hline
              \rowcolor{blue!5} \multicolumn{10}{c}{\textit{Baselines with the Same Experiment Setting}}\\ \hline
              Qwen2.5-VL-7B$^{\star}$ \citep{bai2025qwen2} & \multicolumn{1}{c}{46.6} &48.2&68.2&21.3& 2262.6 &67.1& 59.3 &75.6& 86.0 \\
              \ \ \ \ w/ GRPO \citep{guo2025deepseek} & \multicolumn{1}{c}{54.9} &50.6&68.1&22.5& 2278.6 &60.8& 64.7 &76.5& 85.4 \\
              \ \ \ \ w/ GRPO$^{\dagger}$ \citep{guo2025deepseek} & \multicolumn{1}{c}{53.1} &\underline{51.3}&69.5&18.8& 2167.6 &58.6& 59.1 &76.3& 85.8 \\
              \hline
              \rowcolor{blue!5} \multicolumn{10}{c}{\textit{Our Methods}}\\ \hline
              \ \ \ \ w/ D2D$_{loc}$ (ours) & \multicolumn{1}{c}{54.5} &49.1&68.0&20.0& \underline{2358.4} &65.3& 66.0 &\textbf{77.3}& 87.1 \\
              \ \ \ \ w/ D2I$_{loc}$ (ours) & \multicolumn{1}{c}{\underline{60.6}} &51.1&69.9&26.3& 2267.1 &67.9& 61.1 &76.6& 86.8 \\
              \rowcolor{gray!20}
              \ \ \ \ ${\Delta}_{base}$ & \multicolumn{1}{c}{$+$14.0} & $+$2.9 & $+$1.7 & $+$5.0 & $+$4.5 & $+$0.8 & $+$1.8 & $+$1.0 & $+$0.8 \\
              \rowcolor{gray!20}
              \ \ \ \ ${\Delta}_{grpo\dagger}$ & \multicolumn{1}{c}{$+$7.5} & $-$0.2 & $+$0.4 & $+$7.5 & $+$99.5 & $+$9.3 & $+$2.0 & $+$0.3 & $+$1.0 \\
              \hline
              \ \ \ \ w/ D2D$_{jus}$ (ours) & \multicolumn{1}{c}{28.9} &49.1&65.7&20.1& 2351.1 &\textbf{69.7}& 64.4 &\underline{77.0}& 86.9 \\
              \ \ \ \ w/ D2I$_{jus}$ (ours) & \multicolumn{1}{c}{\textbf{65.0}} &\textbf{53.8}&69.7&19.7& 2299.0 &66.9& 61.4 & 76.4 & 85.8 \\
              \rowcolor{gray!20}\ \ \ \ ${\Delta}_{base}$ & \multicolumn{1}{c}{$+$18.4} & $+$5.6 & $+$1.5 & $-$1.6 & $+$36.4 & $-$0.2 & $+$2.1 & $+$0.8 & $-$0.2 \\
              \rowcolor{gray!20}\ \ \ \ ${\Delta}_{grpo\dagger}$ & \multicolumn{1}{c}{$+$11.9} & $+$2.5 & $+$0.2 & $+$0.9 & $+$131.4 & $+$8.3 & $+$2.3 & $+$0.1 & $+$0.0 \\
              \hline
              \ \ \ \ w/ D2D$_{par}$ (ours) & \multicolumn{1}{c}{52.1} &47.6&62.6&17.6& \textbf{2373.7} &66.4& \underline{67.6} &76.5& \underline{88.8}\\
              \ \ \ \ w/ D2I$_{par}$ (ours) & \multicolumn{1}{c}{60.5} &50.6&\textbf{72.2}&\underline{27.2}& 2219.8 &65.3& 61.6 &76.1& 86.3 \\
              \rowcolor{gray!20}\ \ \ \ ${\Delta}_{base}$ & \multicolumn{1}{c}{$+$13.9} & $+$2.4 & $+$4.0 & $+$5.9 & $-$42.8 & $-$1.8 & $+$2.3 & $+$0.5 & $+$0.3 \\
              \rowcolor{gray!20}\ \ \ \ ${\Delta}_{grpo\dagger}$ & \multicolumn{1}{c}{$+$7.4} & $-$0.7 & $+$2.7 & $+$8.4 & $+$52.2 & $+$6.7 & $+$2.5 & $-$0.2 & $+$0.5 \\
	
            \bottomrule
		\end{tabular}
	}
\caption{Main results of our GEOQA-8K-trained D2I method on in-domain and out-of-domain test sets. \textbf{Bold} and \underline{underline} indicate the best and second-best results, respectively. Qwen w/ GRPO is evaluated with deliberate reasoning, while Qwen w/ GRPO$^{\dagger}$ is evaluated with intuitive reasoning. ${\Delta}{base}$ and ${\Delta}{grpo\dagger}$ denote D2I’s improvements over Qwen2.5-VL-7B and Qwen2.5-VL-7B w/ GRPO$^{\dagger}$, respectively. $^{\star}$ denotes our re-implementation.}
\label{main_results}
\end{table*}

\noindent \textbf{Parsing Consistency (PAR) Strategy.} PAR encourages the model to parse the input image into structured language or symbolic expressions before reasoning. The format reward is assigned when the parsing result appears in the predefined \texttt{<parse>}...\texttt{</parse>} block at the beginning of the response. This promotes global visual understanding as a basis for downstream reasoning. As shown in Fig.~\ref{fig:pipeline}, the expected response format is:
% The PAR strategy supervises the model to output a structured parsing of the input image before generating the reasoning trace. Parsing results, which convert the information in the image into the corresponding structured language or
% symbolic expression, are expected to appear in a predefined format (\ie, a \texttt{<parse>}...\texttt{</parse>} block) at the beginning of the response. The format reward is given when the parsing result follows the correct format. This strategy promotes global visual understanding as a foundation for downstream reasoning. As shown in Fig.~\ref{fig:pipeline}, we expect the model to generate responses in the following format: 

\begin{tcolorbox}
\texttt{<parse>} \textit{parsing result here} \texttt{</parse>} \texttt{<think>} \textit{reasoning here} \texttt{</think>} \texttt{<answer>} \textit{final answer here} \texttt{</answer>}
\end{tcolorbox}
%-----------------------------------------------------------------------

%-----------------------------------------------------------------------
\section{Experiment Settings} \label{sec:5}
%-----------------------------------------------------------------------

%-----------------------------------------------------------------------
\subsection{Datasets}  \label{sec:5.1}
%-----------------------------------------------------------------------

%-----------------------------------------------------------------------
\noindent \textbf{Training.} We train on GEOQA-8K \citep{chen2025r1v}, a geometry-focused multimodal reasoning dataset of text-image pairs requiring mathematical and spatial understanding. The training split contains $8{,}030$ samples.
% We follow the setup of R1V \citep{chen2025r1v} and use GEOQA-8K \citep{chen2025r1v} as our training dataset, which is a geometry-focused multimodal reasoning dataset consisting of text-image pairs that require mathematical and spatial understanding. The training split contains $8030$ examples. 

\noindent \textbf{Evaluation.} We report in-domain performance on the GEOQA-8K test set \citep{chen2025r1v} with $754$ samples. To evaluate generalization, we further test on several out-of-domain benchmarks, including multimodal math benchmarks such as MathVerse \citep{zhang2024mathverse}, MathVista \citep{lu2023mathvista}, and MATH-Vision \citep{wang2024measuring}, as well as general multimodal benchmarks such as MME \citep{fu2024mmecomprehensiveevaluationbenchmark}, MMVet \citep{yu2024mmvetevaluatinglargemultimodal}, MMMU \citep{yue2024mmmu}, SEED-Bench \citep{li2024seed}, and POPE \citep{li2023evaluating}.
% We report the performance on the GEOQA-8K \citep{chen2025r1v} test set as in-domain performance, which includes 754 examples. To assess the generalization of our model, we also evaluate on several widely used out-of-domain benchmarks. Three of them are multimodal math benchmarks: MathVerse \citep{zhang2024mathverse}, MathVista \citep{lu2023mathvista}, and MATH-Vision \citep{wang2024measuring}. These benchmarks encompass a diverse range of mathematical question types, enabling a comprehensive evaluation of the model’s reasoning capabilities. The rest are multimodal benchmarks, including MME \citep{fu2024mmecomprehensiveevaluationbenchmark}, MMVet \citep{yu2024mmvetevaluatinglargemultimodal}, MMMU \citep{yue2024mmmu}, SEED-Bench \citep{li2024seed} and POPE \citep{li2023evaluating}, for general capability evaluation. 
%-----------------------------------------------------------------------

%-----------------------------------------------------------------------
\subsection{Compared Methods} \label{sec:5.2}

We compare D2I with four groups of baselines: closed-source general models, open-source general models, recent reasoning models, and same-setting baselines based on Qwen2.5-VL-7B \citep{bai2025qwen2} with or without GRPO training \citep{guo2025deepseek}. Specifically, these include GPT-4V \citep{gpt4v}, GPT-4o \citep{hurst2024gpt}, Qwen2-VL-7B \citep{bai2023qwen}, InternVL2-8B \citep{chen2024far}, InternVL2.5-8B \citep{chen2024expanding}, LLaVA-CoT-11B \citep{xu2024llava}, R1-Onevision-7B \citep{yang2025r1}, and OpenVLThinker-7B \citep{deng2025openvlthinker}. We further compare D2I with D2D under deliberate reasoning strategies.

More details are shown in Appendices~\ref{ap:implement} and \ref{ap:prompt}.

  \begin{table*}[t]
  \centering
  \tiny

	\resizebox{\linewidth}{!}{
		\begin{tabular}{lcccccccc}
			\toprule
            % \arrayrulecolor{black}
			 \multirow{2}{*}{\bf Method} &  \multicolumn{3}{c}{\textbf{Out-of-domain (Math)}} & \multicolumn{5}{c}{\textbf{Out-of-domain (General)}} \\ 
               \cmidrule(r){2-4} \cmidrule(r){5-9} \noalign{\smallskip} 
             % \hhline{~---------}
             % \noalign{\vskip.2pt}
			& \textbf{\makecell{MathVerse \\ (mini)}} & \textbf{\makecell{MathVista \\ (mini)}} & \textbf{\makecell{MATH-Vision \\ (full)}} & \textbf{\makecell{MME \\ (sum)}} & \textbf{\makecell{MMVet \\ (turbo)}} & \textbf{\makecell{MMMU \\ (val)}} & \textbf{SEED} & \textbf{POPE} \\ \hline

              \rowcolor{blue!5} \multicolumn{9}{c}{\textit{SFT-Only Models}}\\ \hline
              
              SFT$_{GEOQA}$ & 29.1 & 56.8 &14.0 & 2196.6 & 46.2 & 61.6 & 77.1 & 87.9 \\

              SFT$_{loc}$ & 43.2 & 59.6 & 21.8 & 2325.2 &68.1& 64.6 & 75.5 & 85.5 \\

              SFT$_{par}$ & 27.7 &51.0&14.8& 1982.8 &60.8& 66.4 & 77.1 & 86.2 \\
              \hline
              \rowcolor{blue!5} \multicolumn{9}{c}{\textit{SFT-RL Models}}\\ \hline
              
              SFT$-$D2D$_{loc}$ & 46.0 & 56.7 & 17.0 & 2320.7 &61.9& 61.4 & 76.8 & 86.8 \\
              SFT$-$D2I$_{loc}$ & 43.9 & 68.7 & 23.6 & 2318.7 &65.8&64.0 & 75.9 & 81.2 \\
              SFT$-$D2D$_{par}$ & 33.8 & 56.1 & 15.1 & 2353.2 &46.8& 66.3 & 75.2 & 88.3 \\
              SFT$-$D2I$_{par}$ & 36.6 & 67.0 & 20.8 & 2302.5 &60.6& 65.8 & 76.9 & 86.4 \\

              \hline
              \rowcolor{blue!5} \multicolumn{9}{c}{\textit{RL-Only Models}}\\ \hline
              D2D$_{loc}$ (ours) &49.1&68.0&20.0& 2358.4 &65.3& 66.0 &\textbf{77.3}& 87.1 \\
              D2I$_{loc}$ (ours) &51.1&69.9&26.3& 2267.1 &67.9& 61.1 &76.6& 86.8 \\
              % \hline
              D2D$_{jus}$ (ours) &49.1&65.7&20.1& 2351.1 &\textbf{69.7}& 64.4 &77.0& 86.9 \\
              D2I$_{jus}$ (ours) &\textbf{53.8}&69.7&19.7& 2299.0 &66.9& 61.4 & 76.4 & 85.8 \\
              % \hline
              D2D$_{par}$ (ours) &47.6&62.6&17.6& \textbf{2373.7} &66.4& \textbf{67.6} &76.5& \textbf{88.8}\\
              D2I$_{par}$ (ours) &50.6&\textbf{72.2}&\textbf{27.2}& 2219.8 &65.3& 61.6 &76.1& 86.3 \\
	
            \bottomrule
      \end{tabular}
      }
\caption{Comparison of our method (RL-Only) with SFT-only and SFT-RL models. \textbf{Bold} denotes the best results.}
  \label{ablation}
  \end{table*}

% The methods include four aspects: (1) closed-source general models (\ie, GPT-4V \citep{gpt4v} and GPT-4o \citep{hurst2024gpt}), (3) open-source general models (\ie, Qwen2-VL-7B \citep{bai2023qwen}, InternVL2-8B \citep{chen2024far} and InternVL2.5-8B \citep{chen2024expanding}), (3) recent reasoning models (\ie, LLaVA-CoT-11B \citep{xu2024llava}, R1-Onevision-7B \citep{yang2025r1} and OpenVLThinker-7B \citep{deng2025openvlthinker}), and (4) baselines with the same experiment setting (\ie, Qwen2.5-VL-7B \citep{bai2025qwen2} itself and with GRPO training \citep{guo2025deepseek}). We also compare D2I with D2D in our designed deliberate reasoning strategies.
%-----------------------------------------------------------------------

%-----------------------------------------------------------------------
\section{Experiental Results and Analysis} \label{sec:6}
%-----------------------------------------------------------------------
% 加入RQ的形式，比如第一个问题How does D2I’s quantitative performance compare to competitive baselines? --》 main results (RQ1)
% 第二个问题是d2i相比d2d的提升是否源自于模型没有相应的d reasoning能力。对应ablation study，sft only和sft-rl的对比 --> ablation study (RQ2)
% 问题三：How does the qualitative performance of D2I stand against competitive baselines? --> case study (RQ3)
% 第4个问题是探索为什么d2i有更好的性能，它到底改变了什么？More Exploration (RQ4) 分三个小段，展示pass@k曲线（放不下的放附录），token shift比例和词云图（具体的例子放一个在正文，多的放附录），entropy柱状图对比（画箱线图，如果效果不好就画平均值的柱状图）。
% 还有一个问题，在非数学的训练集上还有效吗 --〉results on doc-mix (appendix) 做了主实验，sft消融。在前面实验设置里把doc相关内容删掉
%-----------------------------------------------------------------------
Our experimental analysis addresses the following \textit{Research Questions (RQs)}:
% We organize our experimental analysis around these following \textit{Research Questions (RQs)}:

\begin{itemize}
    \item \textit{\textbf{RQ1}: How does D2I with different deliberate reasoning strategies compare with competitive baselines quantitatively?}
    \item \textit{\textbf{RQ2}: Does D2I outperform D2D because D2D lacks deductive reasoning ability?}
    \item \textit{\textbf{RQ3}: How do the three strategies, D2I, and D2D affect model behavior?}
    \item \textit{\textbf{RQ4}: How does D2I compare with competitive baselines qualitatively?}

    % \item \textit{\textbf{RQ1}: How does quantitative performance of D2I trained with different deliberate reasoning strategies compare to competitive baselines?}
    % \item \textit{\textbf{RQ2}: Is D2I’s improvement over D2D due to the model’s lack of deductive reasoning capability?}
    % \item \textit{\textbf{RQ3}: How does the qualitative performance of D2I stand against competitive baselines?}
    % \item \textit{\textbf{RQ4}: What exactly does designed three atrategies, D2I and D2D influence in the models?}
    % \item \textit{\textbf{RQ5}: Is D2I still effective on non-math training sets?}
\end{itemize}

%-----------------------------------------------------------------------
\subsection{Main Results (\textit{\textbf{RQ1}})}  \label{sec:6.1}
%-----------------------------------------------------------------------
\begin{figure*}[t]
	\centering
        \includegraphics[width=\textwidth]{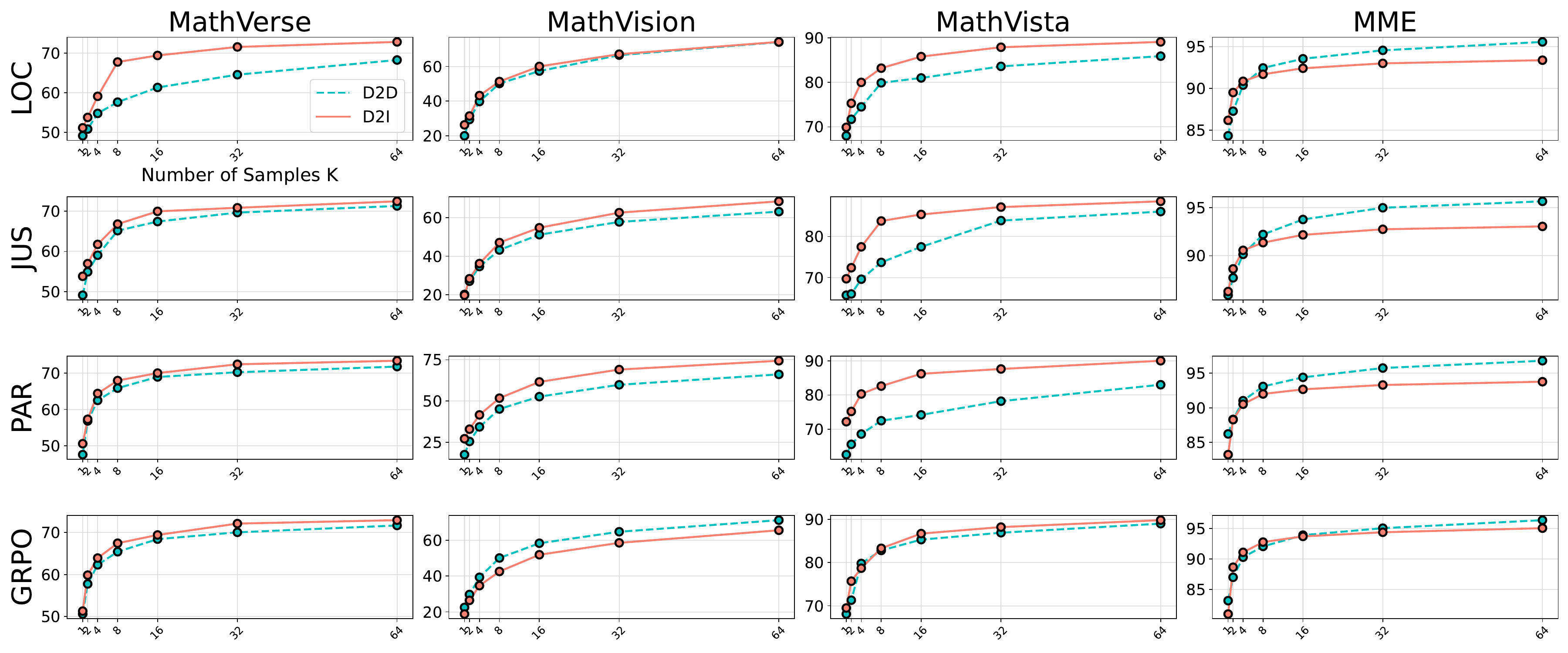}
	\caption{Pass@$k$ results on math benchamrks and general benchmarks.}
	\label{fig:passk}
\end{figure*}

\begin{figure*}[t]
	\centering
        \includegraphics[width=\textwidth]{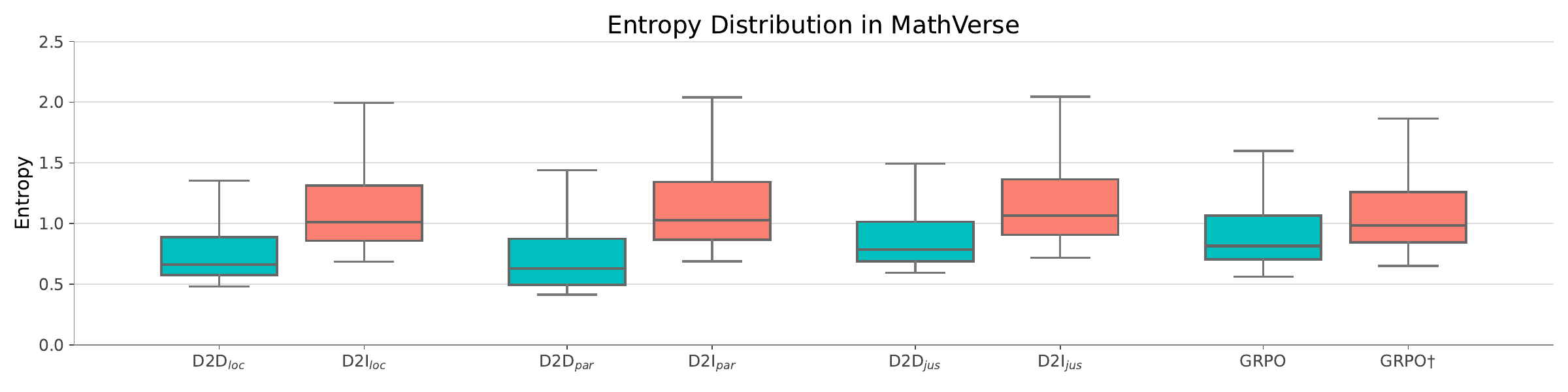}
	\caption{Entropy distribution results on MathVerse dataset.}
	\label{fig:entropy}
\end{figure*}
%-----------------------------------------------------------------------

The results of training on the mathematical dataset are summarized in Table~\ref{main_results}. D2I and D2D achieve strong performance on both in-domain and out-of-domain benchmarks. D2I consistently outperforms the Qwen2.5-VL-7B base model and the GRPO baseline on nearly all benchmarks. On the in-domain math benchmark, D2I improves performance by at least $13.9\%$ over the base model and $7.4\%$ over GRPO. On out-of-domain math benchmarks, D2I$_{loc}$ and D2I$_{par}$ obtain gains of $1\%$--$8\%$, with slight drops on MathVerse, while D2I$_{jus}$ achieves gains of $1\%$--$6\%$, with a slight drop on MATH-Vision.

Although D2I does not surpass GPT-4o on a few general benchmarks, such as MATH-Vision and MMMU, it consistently outperforms other open-source general and reasoning models. These results demonstrate the effectiveness of our D2I framework and deliberate reasoning strategies for multimodal reasoning.

We attribute these improvements to two factors. First, format-constrained training promotes structured and interpretable reasoning patterns. Second, unconstrained inference allows more flexible generation and a larger response search space, increasing the chance of reaching the correct answer without being forced to produce potentially low-quality deliberate outputs, such as boxes or parsing results. This is particularly beneficial for math benchmarks, although it may sometimes introduce overly divergent reasoning on general benchmarks and negatively affect performance. We further analyze the differences among the three strategies in Appendix~\ref{ap:differentstrategy}, verify the effect of non-math training set in~\ref{ap:doc}, discuss other reasoning methods in~\ref{ap:discuss}, compare different model scale in~\ref{ap:model_size} and different backbone in~\ref{ap:backbone}.

\subsection{Ablation Study \textit{(\textbf{RQ2})}}  \label{sec:6.2}
% 改成sft only的模型和sft-rl的模型
% 首先，我们的方法在sft模型中也有优势。其次，sft-rl的D2D于D2I的大小关系与RL-only的大部分相同，，说明RL-only中D2D没有D2I效果好是不是由于没有到生成box和parse的能力。并且sft-rl的D2D并没有比rl-only的D2D结果好，说明SFT在我们的架构中不是影响模型表现的关键因素。
%-----------------------------------------------------------------------

To test whether D2I’s improvement over D2D is caused by D2D’s limited deductive reasoning ability, we compare RL-only models with SFT-only and SFT-RL variants. Training details are provided in Appendix~\ref{ap:sft}. As shown in Table~\ref{ablation}, RL-only models consistently outperform both SFT-only and SFT-RL baselines on most benchmarks under both LOC and PAR strategies. In addition, D2I generally outperforms the corresponding D2D variant in both RL-only and SFT-RL settings.
These results indicate that the D2I--D2D gap is not merely due to D2D’s difficulty in learning accurate box or parsing outputs. Rather, RL plays a central role in improving reasoning capability, while SFT is not the main performance driver. This is further supported by SFT$_{GEOQA}$, which is trained on the same dataset as our main setting but still underperforms our RL-only models.

\subsection{More Exploration \textit{(\textbf{RQ3})}}  \label{sec:6.3}

\noindent \textbf{Reasoning Upper Bound.} We further analyze Pass@$k$ \citep{chen2021evaluating}, as shown in Fig.~\ref{fig:passk}. D2I consistently outperforms D2D across most benchmarks, particularly at smaller $k$. This indicates that D2I enables the model to generate more accurate responses in higher-ranked hypotheses, suggesting better guidance during inference and a higher upper bound. 
% indicating better ranking of correct responses and higher sample efficiency. 
While the D2I--D2D gap in the GRPO baseline narrows as $k$ increases, the gap remains larger under our deliberate reasoning strategies. This suggests that these strategies more strongly shape model behavior and help D2I reach correct answers with fewer samples.

% we analyze the Pass@$k$ \citep{chen2021evaluating} as shown in Fig. \ref{fig:passk}. Across most benchmarks, D2I consistently outperforms D2D, particularly at smaller values of $k$. This indicates that D2I enables the model to generate more accurate responses in higher-ranked hypotheses, suggesting better guidance during inference and a higher upper bound. 
% Moreover, we observe that in the GRPO baseline, the performance of D2D and D2I tends to converge as $k$ increases. In contrast, under our three deliberate reasoning strategies, the gap between D2D and D2I remains more pronounced. This indicates that deliberate reasoning strategies exert a stronger enhancing or suppressing effect on D2I, demonstrating their greater influence in shaping model behavior. It also highlights the effectiveness of deliberate reasoning strategies in improving sample efficiency, as D2I is more likely to hit the correct answer with fewer samples. 

% On the general benchmark (\ie MME), at larger $k$, D2D outperforms D2I and the trend reverses. This suggests that while D2I benefits from a more aggressive or diverse output policy on math tasks, it may lead to a limited upper bound on performance in simpler, general reasoning scenarios.

\noindent \textbf{Entropy Distribution.} We further analyze output distribution entropy \citep{cheng2025reasoning} to examine model exploration. Entropy serves as a proxy for how deterministic or exploratory a model's generation policy is: lower entropy implies more deterministic outputs, while higher entropy suggests a broader sampling space and increased diversity. We visualize the entropy distribution of MathVerse as shown in Fig.~\ref{fig:entropy}, D2I shows a higher and broader entropy distribution than D2D, indicating a more diverse and less deterministic generation policy. Additional results are provided in Appendix~\ref{ap:entropy}.

% To further examine the exploration characteristics of our models, we conduct entropy-based analysis, measuring the output distribution entropy \citep{cheng2025reasoning} for each model under different prompting conditions. Entropy serves as a proxy for how deterministic or exploratory a model's generation policy is: lower entropy implies more deterministic outputs, while higher entropy suggests a broader sampling space and increased diversity. We visualize the entropy distribution using box plots, comparing D2D and D2I across benchmarks as shown in Fig. \ref{fig:entropy}. The results reveal that D2I exhibits a broader and higher entropy distribution than D2D, confirming that D2I encourages more diverse and less deterministic generation patterns. We also show the entropy distribution on other benchmarks in Appendix~\ref{ap:entropy}. 

\begin{figure}
	\centering
    \includegraphics[width=0.5\textwidth]{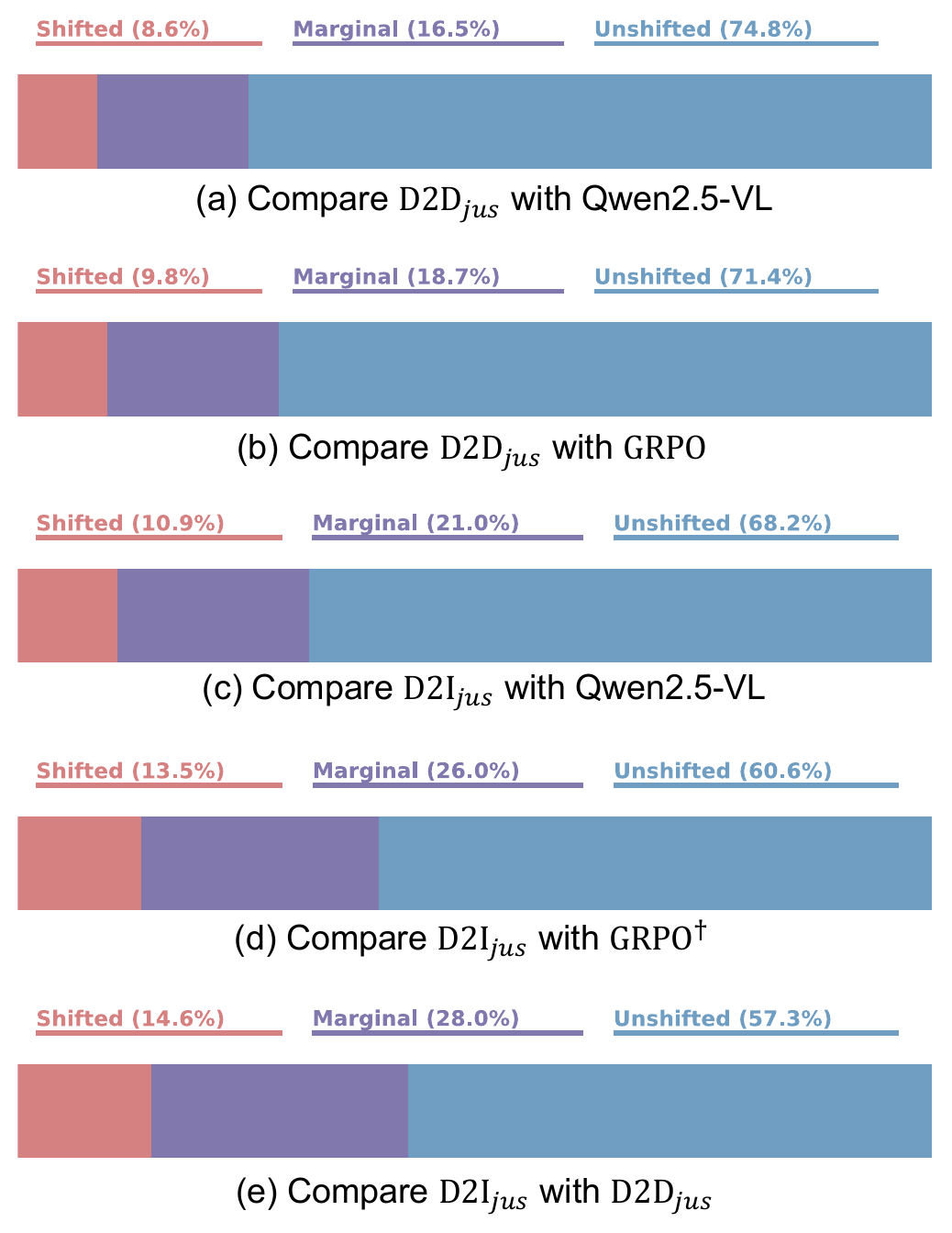}
	\caption{Results of token distribution shift for JUS strategy on MATH-Vision.}
	\label{fig:tokenshift}
\end{figure}

\noindent \textbf{Token Distribution Shift.} To examine inference-time behavior, we analyze token-level output shifts under the same input prompt \citep{lin2023unlocking}. We compare responses from our deliberate reasoning strategies with those from GRPO-trained and base models to measure how strongly each model changes its output distribution.
As shown in Fig.~\ref{fig:tokenshift}, shifted, marginal, and unshifted tokens correspond to selected-token rank differences of $\geq 3$, $1$--$2$, and $0$, respectively. D2I consistently produces a higher shifted-token ratio than D2D, suggesting a more exploratory generation pattern and a stronger restructuring of the output space. This helps D2I explore alternative solutions, especially on math benchmarks. Additional results and word-cloud visualizations are provided in Appendices~\ref{ap:tokenshift} and~\ref{ap:cloud}.

% To investigate the 
% behavioral differences between models during inference, we analyze how each model modifies its token-level output given the same input prompt \citep{lin2023unlocking}. 
% Specifically, we compare the response tokens of our deliberate reasoning strategy with those of baseline models, including GRPO-trained and base model variants. This comparison quantifies how aggressively each model reshapes its output distribution.
% As shown in Fig. \ref{fig:tokenshift}, shifted tokens refer to token positions where the selected token ranks differ by three or more between the two models, marginal tokens indicate a rank difference of one to two, and unshifted tokens represent positions where the two models select the same token. D2I models consistently introduce a higher proportion of shifted tokens relative to their D2D counterparts, indicating a more exploratory generation pattern. This pattern of token-level shift implies that D2I is not merely fine-tuning or adjusting local generation behavior but is actively restructuring the output space. The higher token shift ratio of D2I reflects its capacity to escape suboptimal local minima and explore alternative outputs, particularly beneficial in math benchmarks. We also show the token distribution shift results on other experimental settings in Appendix~\ref{ap:tokenshift} and word cloud visualizations to more intuitively illustrate which specific tokens were altered in Appendix~\ref{ap:cloud}.

%-----------------------------------------------------------------------
\subsection{Case Study \textit{(\textbf{RQ4})}}  \label{sec:6.4}

\begin{figure}[t!]
    \begin{center}
        \includegraphics[width=0.49\textwidth]{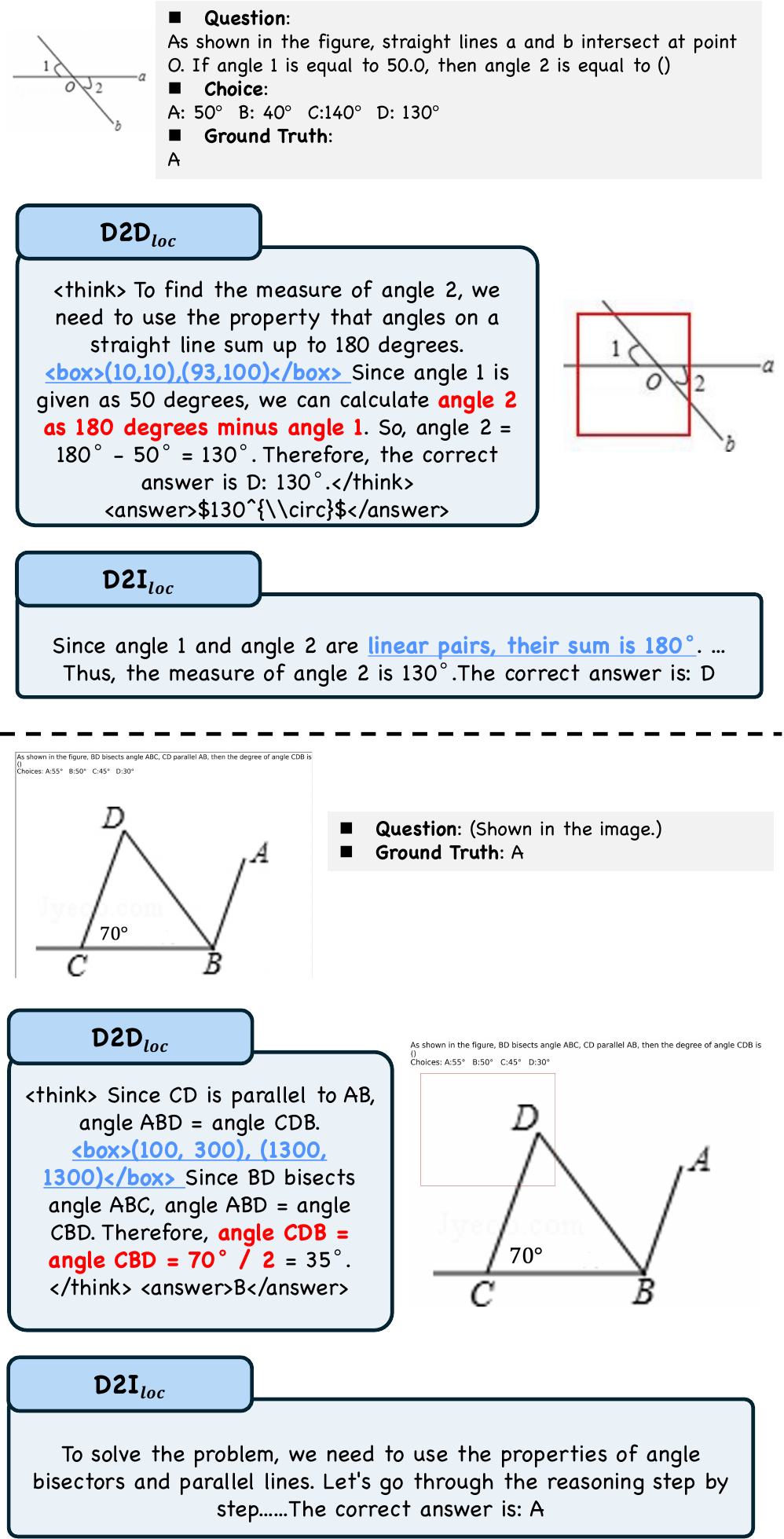}
    \end{center}
	\caption{Failed cases under the D2D paradigm. The red box beside the D2D$_{loc}$ response visualizes the region indicated by the coordinates generated in \texttt{<box></box>}.}
	\label{fig:more_case}
\end{figure}

Analyzing D2D failures is also important. As shown in Fig.~\ref{fig:more_case}, we find several cases, especially under LOC strategy, where model localizes a plausible crucial region but still predicts the wrong answer. This suggests that explicit localization does not necessarily translate into correct reasoning.

We attribute this to three limitations of D2D. First, a bounding box only guides attention, but does not guarantee correct semantic interpretation or integration into the reasoning process. Second, D2D conditions later reasoning on its own generated deliberate steps, so inaccurate boxes or parsing results can propagate errors through the reasoning chain. Third, enforcing explicit structures such as \texttt{<box>}...\texttt{</box>} or \texttt{<parse>}...\texttt{</parse>} at test time restricts exploration and makes the model overly dependent on potentially suboptimal intermediate artifacts. By removing these explicit constraints during inference, D2I can leverage the visual understanding learned during training while maintaining more flexible reasoning paths. More cases can be found in Appendix~\ref{ap:more_case}.

% \begin{figure}[h]
% 	\centering
%     \includegraphics[width=0.5\linewidth]{case_2.pdf}
% 	\caption{Inference responses to the same math question from models trained with different deliberate reasoning strategies.}
% 	\label{fig:curve}
% \end{figure}

% %-----------------------------------------------------------------------
% We analyze a representative failure case where GRPO, D2I$_{loc}$, and D2I$_{par}$ all produce incorrect answers, while D2I$_{jus}$ succeeds. As shown in Fig. \ref{fig:curve}, notably, all four responses share a highly similar structure and format, demonstrating the same reasoning steps. 
% However, only the D2I$_{jus}$ model correctly identifies a critical concept in its reasoning trace: the mention of \textit{vertical angles}. This small but crucial difference allows D2I$_{jus}$ to reach the correct final answer, whereas the other models fail. 
% This highlights an important observation: \textbf{Once the key visual perceptions are correctly understood, the model can form a complete and accurate reasoning chain, with no explicit output format or structured intermediate representation required}. 
% It underscores that understanding and articulating the right concept, rather than strictly enforcing output formats, is often the deciding factor in successful reasoning. 
% This case also illustrates the unique strength of D2I$_{jus}$ in aligning with the LLM's natural generation behavior. 

%-----------------------------------------------------------------------

%-----------------------------------------------------------------------
\section{Conclusion}
%-----------------------------------------------------------------------

In this work, we propose the Deliberate-to-Intuitive reasoning framework for improving MLLMs on complex visual reasoning tasks. Motivated by the gap between training-time supervision and inference-time behavior, D2I encourages models to engage in deeper reasoning during training through format-constrained, goal-driven reinforcement learning, while allowing flexible and intuitive generation at test time.
We design three types of deliberate reasoning strategies to guide the model toward stronger semantic understanding and reasoning capability. Extensive experiments on both in-domain and out-of-domain benchmarks demonstrate that D2I outperforms strong baselines across a wide range of reasoning challenges. Notably, D2I shows competitive or superior performance even compared to some proprietary models, despite relying only on lightweight rule-based rewards and no additional annotation.
We believe that D2I offers a scalable, annotation-free strategy for enhancing the reasoning abilities of MLLMs. In future work, we plan to explore further applications to other multimodal domains such as science diagrams, instructional videos, or procedural planning.

\section*{Limitations}

This work has several limitations. First, D2I depends on manually designed deliberate reasoning strategies and rule-based format rewards. While these designs avoid additional human annotations, they may not cover all types of multimodal reasoning.
On the other hand, our format reward mainly checks whether the response follows the expected structure, such as producing the required tags or fields. However, it does not directly evaluate the semantic quality or correctness of the intermediate reasoning steps. As a result, the model may satisfy the format constraint even when its localized region, visual justification, or parsed representation is imperfect. Future work should explore content-aware rewards that assess not only whether the response follows the format, but also whether the intermediate reasoning is semantically correct and faithfully grounded in the visual input.

% \section*{Acknowledgments}

% This document has been adapted
% by Steven Bethard, Ryan Cotterell and Rui Yan
% from the instructions for earlier ACL and NAACL proceedings, including those for
% ACL 2019 by Douwe Kiela and Ivan Vuli\'{c},
% NAACL 2019 by Stephanie Lukin and Alla Roskovskaya,
% ACL 2018 by Shay Cohen, Kevin Gimpel, and Wei Lu,
% NAACL 2018 by Margaret Mitchell and Stephanie Lukin,
% Bib\TeX{} suggestions for (NA)ACL 2017/2018 from Jason Eisner,
% ACL 2017 by Dan Gildea and Min-Yen Kan,
% NAACL 2017 by Margaret Mitchell,
% ACL 2012 by Maggie Li and Michael White,
% ACL 2010 by Jing-Shin Chang and Philipp Koehn,
% ACL 2008 by Johanna D. Moore, Simone Teufel, James Allan, and Sadaoki Furui,
% ACL 2005 by Hwee Tou Ng and Kemal Oflazer,
% ACL 2002 by Eugene Charniak and Dekang Lin,
% and earlier ACL and EACL formats written by several people, including
% John Chen, Henry S. Thompson and Donald Walker.
% Additional elements were taken from the formatting instructions of the \emph{International Joint Conference on Artificial Intelligence} and the \emph{Conference on Computer Vision and Pattern Recognition}.

\bibliography{main}

\appendix

% \section{The Use of Large Language Models}

% In the preparation of this manuscript, LLMs were used solely for the purpose of text polishing, including grammar correction and stylistic refinement. No content was generated or rewritten with the intention of altering the scientific meaning, originality, or conclusions of the work. All ideas, analyses, and results presented in this paper are entirely the authors’ own.

%-----------------------------------------------------------------------
\section{Implementation} \label{ap:implement}

We adopt Qwen2.5-VL-7B-Instruct \citep{bai2025qwen2} as our base model for all experiments. We train the model for 150 steps with the batch size of $128$, the learning rate of $1e-6$, the max response length of $1024$ tokens, and the sampling temperature of $1$. \textcolor{black}{The inference-time hyperparameters for any setting are kept consistent with those used during training.} During training and evaluation, we utilize the prompt as shown in Appendix~\ref{ap:prompt}. In D2I, we train the model with the designed deliberate reasoning strategies by prompts shown in Training Stage of Appendix~\ref{ap:prompt}, and evaluate it with the intuitive reasoning style by prompts shown in Inference Stage of Appendix~\ref{ap:prompt}. In D2D, during both training and evaluation, we use the same prompts with the designed deliberate reasoning strategies by prompts shown in Training Stage of Appendix~\ref{ap:prompt}. All experiments are conducted on $8$ NVIDIA A100 GPUs, each with $80$GB of memory.
%-----------------------------------------------------------------------

\section{Prompts}
%-----------------------------------------------------------------------
\label{ap:prompt}
%-----------------------------------------------------------------------
\noindent \textbf{Training Stage.} The prompt used in the LOC strategy is as follows:

\begin{tcolorbox}
You are a helpful assistant. First think about the reasoning process for answering the question. As part of your reasoning, identify the crucial region of the image needed to answer the question. Before reaching your final conclusion in the reasoning, output the coordinate of this crucial region. Your response should follow this structure. The reasoning process, the coordinate of the crucial region for answering this question, and answer are enclosed within \texttt{<think> </think>}, \texttt{<box> </box>}, and \texttt{<answer> </answer>} tags, respectively, i.e., \texttt{<think>} Your step-by-step reasoning process... \texttt{<box>} (x1, y1), (x2, y2) \texttt{</box>} the following step-by-step reasoning process based on this region... \texttt{</think><answer>} answer here \texttt{</answer>}. 
\end{tcolorbox}

The prompt used in the JUS strategy is as follows:

\begin{tcolorbox}
You are a helpful assistant. User asks questions, Assistant solves step by step. First think about reasoning process within \texttt{<think></think>} tags. In addition to the reasoning steps leading to your answer, identify the crucial part(s) of the image that are essential for solving the question. Describe why these parts are important and explain how they contribute to your reasoning within \texttt{<crucial></crucial>}. Final answer goes in \texttt{<answer> </answer>} tags. The overall response format is \texttt{<think>} reasoning here \texttt{</think> <crucial>} crucial part description here \texttt{</crucial> <answer>}final answer here \texttt{</answer>}.
\end{tcolorbox}

The prompt used in the PAR strategy is as follows:

\begin{tcolorbox}
You are a helpful assistant. You FIRST parse image with structure language, and then think about reasoning process as an internal monologue and finally provide final answer. The parsing result MUST BE enclosed within \texttt{<parse> </parse>} tags. The reasoning process MUST BE enclosed within \texttt{<think> </think>} tags. The final answer MUST BE enclosed within \texttt{<answer> </answer>} tags. Note the parsing result is parsing image with listing all the objects and their relationships using predicate format. The overall response format is \texttt{<parse>} parsing result here \texttt{</parse> <think>} reasoning process here \texttt{</think> <answer>} final answer here \texttt{</answer>}.
\end{tcolorbox}

The prompt used in the GRPO training is as follows:

\begin{tcolorbox}[breakable]
A conversation between User and Assistant. The user asks a question, and the Assistant solves it. The assistant first thinks about reasoning process in the mind and then provides the user with answer. The reasoning process and answer are enclosed within \texttt{<think> </think>} and \texttt{<answer> </answer>} tags, respectively, i.e., \texttt{<think>} reasoning process here \texttt{</think> <answer>} answer here \texttt{</answer>}.
\end{tcolorbox}

\noindent \textbf{Inference Stage.} The prompt used in the D2I framework is as follows:

\begin{tcolorbox}
You are a helpful assistant. Please provide step-by-step reasoning process first, and then provide your final answer.
\end{tcolorbox}
%-----------------------------------------------------------------------

\section{Comparison of Different Deliberate Reasoning Strategies} 

\label{ap:differentstrategy}

As shown in Table \ref{main_results}, among the three deliberate reasoning strategies, we observe that their performance varies across benchmarks, with no single strategy consistently dominating the others. Instead, each strategy shows strengths on specific types of tasks, reflecting complementary reasoning capabilities. The JUS strategy tends to perform better on benchmarks such as MathVerse and MMVet, which are explicitly designed to test whether MLLMs truly interpret math diagrams and demand expert-level multimodal understanding and reasoning. In both cases, \textbf{JUS’s natural-language explanations help align the model’s reasoning process} with the expectations of these benchmarks, making it easier to leverage visual cues in a text-friendly format. The LOC strategy shows strong results on MME and SEED, emphasizing accurate spatial grounding and fine-grained perception. By learning to identify bounding box coordinates of crucial regions, \textbf{LOC strategy enables the model to anchor its reasoning in the correct image region}, improving reliability in tasks sensitive to spatial detail. The PAR strategy performs well on MathVista, MATH-Vision, MMMU, and other benchmarks, which often involve complex visual layouts and structural relationships. For example, MathVista includes diverse visuals (\eg, puzzle pieces, plots, and scientific diagrams), necessitating deep structural understanding. \textbf{The PAR strategy’s emphasis on modeling global object structure helps the model form a coherent representation of the image}, which is critical for such benchmarks.

%-----------------------------------------------------------------------

\section{Results on Non-Math Training Set}  \label{ap:doc}
%-----------------------------------------------------------------------

\begin{table*}[t]
% \belowrulesep=0pt
% \aboverulesep=0pt
% \renewcommand\arraystretch{1.15}
\centering

	\resizebox{\linewidth}{!}{
		\begin{tabular}{lccccccccc}
			\toprule
            % \arrayrulecolor{black}
			 \multirow{2}{*}{\bf Method} & \textbf{In-domain} & \multicolumn{3}{c}{\textbf{Out-of-domain (Math)}} & \multicolumn{5}{c}{\textbf{Out-of-domain (General)}} \\ 
               \cmidrule(r){2-2} \cmidrule(r){3-5} \cmidrule(r){6-10} \noalign{\smallskip} 
             % \hhline{~---------}
             % \noalign{\vskip.2pt}
			  & \textbf{Doc-Mix} & \textbf{\makecell{MathVerse \\ (mini)}} & \textbf{\makecell{MathVista \\ (mini)}} & \textbf{\makecell{MATH-Vision \\ (full)}} & \textbf{\makecell{MME \\ (sum)}} & \textbf{\makecell{MMVet \\ (turbo)}} & \textbf{\makecell{MMMU \\ (val)}} & \textbf{SEED} & \textbf{POPE} \\ \hline

              \rowcolor{blue!5} \multicolumn{10}{c}{\textit{Closed-Source General Models}}\\ \hline

              GPT-4V \citep{gpt4v}& -- & 39.4 & 58.1 & 22.7 & 1926.6 & 67.5 & 63.1 & 53.8 & -- \\
              GPT-4o \citep{hurst2024gpt}& -- & \textbf{50.2} & 63.8 &\textbf{30.3} & -- & \underline{69.1} & \textbf{69.1} & 72.0 & 86.9 \\ \hline
              \rowcolor{blue!5} \multicolumn{10}{c}{\textit{Open-Source General Models}}\\ \hline
              Qwen2-VL-7B \citep{bai2023qwen}& -- & 31.9 & 58.2 & 16.3 & 2326.8 & 62.0 & 54.1 & 75.1 & 88.1 \\
              InternVL2-8B \citep{chen2024far}& -- & 37.0 &58.3&18.4&2210.3&54.2&52.6 & --& 86.9 \\
              InternVL2.5-8B \citep{chen2024expanding}& -- & 39.5 & 64.4 & 19.7 & 2344.1 & 62.8& 56.0 & -- & \textbf{90.6}  \\

              \hline
              \rowcolor{blue!5} \multicolumn{10}{c}{\textit{Reasoning Models}}\\ \hline
              LLaVA-CoT-11B \citep{xu2024llava}& -- & 20.3  & 54.8 & -- & -- & 60.3 & -- & -- & --  \\
              
              R1-Onevision-7B \citep{yang2025r1}& -- & 46.4 & 64.1 & -- & 2192.2 & 67.5 & -- &  66.5 & 84.9 \\
              OpenVLThinker-7B \citep{deng2025openvlthinker}& -- & 47.9 & \textbf{70.2} & \underline{25.3} & --& --& --& --& -- \\
              % MM-Eureka-7B & \multicolumn{1}{c}{--} & 50.3 & 73.0 & -- & --& --& --& --& -- \\
              \hline
              \rowcolor{blue!5} \multicolumn{10}{c}{\textit{Baselines with the Same Experiment Setting}}\\ \hline
              Qwen2.5-VL-7B$^{\star}$ \citep{bai2025qwen2}&78.3 &48.2&68.2&21.3&2262.6&67.1&59.3&75.6&86.0 \\
              \ \ \ \ w/ GRPO  \citep{guo2025deepseek} & 81.1&47.4&67.9&23.0 &\textbf{2382.1} &66.1&65.8&\textbf{77.6}&\underline{88.7} \\
              \ \ \ \ w/ GRPO$^{\dagger}$ \citep{guo2025deepseek}&78.8 &43.3&65.4&19.1 & 2292.5 &66.8&61.1&76.1&86.3 \\
              \hline
              \rowcolor{blue!5} \multicolumn{10}{c}{\textit{Our Methods}}\\ \hline
              \ \ \ \ w/ D2D$_{loc}$ (ours) & \underline{81.5} & 42.2 &64.8&20.3 &\underline{2363.6}&67.8& \underline{67.2} & \textbf{77.6} & 87.8 \\
              \ \ \ \ w/ D2I$_{loc}$ (ours) &80.7&\underline{48.8}&68.9&22.2 & 2311.3 &\textbf{69.8}& 62.0 & \underline{76.7} & 86.3 \\
              \rowcolor{gray!20}
              \ \ \ \ ${\Delta}_{base}$&$+$2.4 & $+$0.6 & $+$0.7 & $+$0.9 & $+$48.7 & $+$2.7 & $+$2.7 & $+$1.1 & $+$0.3 \\
              \rowcolor{gray!20}
              \ \ \ \ ${\Delta}_{grpo\dagger}$&$+$1.9 & $+$5.5 & $+$3.5 & $+$3.1 & $+$18.8 & $+$3.0 & $+$0.9 &$+$0.6 &$+$0.0 \\
              \hline
              \ \ \ \ w/ D2D$_{jus}$ (ours)&\textbf{82.0} &45.8 &64.7&19.7&2357.4 &66.8& 65.3 & \underline{76.7} & 88.2  \\
              \ \ \ \ w/ D2I$_{jus}$ (ours)&78.7 & 48.6 & \underline{70.1} & 21.3 & 2286.7 &68.6& 59.6 & 76.3 & 86.8  \\
              \rowcolor{gray!20}\ \ \ \ ${\Delta}_{base}$ & $+$0.4 & $+$0.4 & $+$1.9 & $+$0.0 & $+$24.1 & $+$1.5 &$+$0.3 & $+$0.7 & $+$0.8 \\
              \rowcolor{gray!20}\ \ \ \ ${\Delta}_{grpo\dagger}$ &$-$0.1 & $+$5.3 & $+$4.7 & $+$2.2 & $-$5.8 &$+$1.8 & $-$1.5 & $+$0.2 & $+$0.5 \\

            \bottomrule
		\end{tabular}
	}
\caption{Main results of our proposed D2I models trained on self-constructed non-math dataset. The \textbf{bold} denotes the best performance, while the \underline{underline} denotes the second best one. In the Qwen w/ GRPO, we evaluate the model with the deliberate reasoning, while in w/ GRPO$^{\dagger}$ we evaluate with the intuitive reasoning. ${\Delta}_{base}$ represents the improvement of D2I over Qwen2.5-VL-7B, while ${\Delta}_{grpo\dagger}$ represents the improvement over Qwen2.5-VL-7B w/ GRPO$^{\dagger}$. $^{\star}$ denotes our re-implementation.}
\label{main_results_doc}
\end{table*}

To further verify whether the results are consistent on datasets beyond the mathematical domain, we constructed a mixed dataset in the document VQA field
% , named Doc-Mix\footnote{\url{https://huggingface.co/datasets/YahanYu/Doc-Mix}}
, which includes samples from DocVQA \citep{mathew2021docvqa}, InfographicVQA \citep{mathew2022infographicvqa}, ArxivQA \citep{li2024multimodal}, and TAT-DQA \citep{zhu2022towards,zhu2024doc2soargraph}. The training set contains 8,040 examples. For inference, we report the performance on the Doc-Mix test set as in-domain performance, which includes 720 examples. 
We also evaluate on the widely used out-of-domain benchmarks: MathVerse \citep{zhang2024mathverse}, MathVista \citep{lu2023mathvista}, and MATH-Vision \citep{wang2024measuring}.
Results are summarized in Table~\ref{main_results_doc}. Compared to other methods, the overall trend is not as strong as D2I trained on GEOQA-8K, since the model trained on document VQA data mostly achieved only the second-best performance. However, this is reasonable because we only used document VQA datasets and did not see any math-related samples during training. Nevertheless, our approach consistently outperforms GRPO and Qwen2.5-VL, demonstrating that our method maintains consistent effectiveness across different training data.
%-----------------------------------------------------------------------

\section{Discussion with Other Reasoning Methods}  \label{ap:discuss}
%-----------------------------------------------------------------------

\textcolor{black}{Recently, research on reasoning has gained significant traction. The following distinguished works have captured our attention, and we proceed here with a comparative analysis against our proposed framework:}

\begin{enumerate}
    \item \textcolor{black}{Vision-R1 \citep{huang2025vision} introduced the core R1 paradigm that uses rule-based rewards to incentivize MLLMs to generate structured reasoning paths. It proved that format constraints can effectively enforce deliberate behavior and teach new skills without human content supervision. But the rigid requirement for structured output at inference time severely suppresses exploration and introduces brittleness, leading to suboptimal performance on OOD tasks compared to D2I.}
    \item \textcolor{black}{R1-VL \citep{zhang2025r1} refined the R1 framework by introducing GRPO, focusing on step-wise relative quality feedback to optimize the policy. It provided the highly efficient and stable GRPO that D2I's deliberate training phase directly utilizes, significantly accelerating skill acquisition. But it is a coupled D2D model. The enhanced reasoning skill learned efficiently through GRPO is still constrained by the format requirement at inference. D2I unlocks this skill by removing the constraint.}
    \item \textcolor{black}{Video-R1 \citep{feng2025video} applied the R1 paradigm to video reasoning tasks, designing temporal-specific rule rewards to enhance multi-frame, deliberate processing in MLLMs. It validated the cross-modal generalizability of the R1 concept to complex sequence modalities like video. However, video reasoning is resource-intensive. Mandating structured D2D output exacerbates latency and inflexibility. D2I's principle of intuitive (concise) inference is even more crucial here to maintain efficiency and generalizability.}
    \item \textcolor{black}{Seg-Zero \citep{liu2025seg} utilized R1-style reinforcement to guide and improve non-linguistic, structured visual output via a reasoning chain. It demonstrated that format reinforcement can drive the generation of structured outputs that are not pure text, linking thought chains to low-level visual perception tasks. But its D2D nature is necessary for the task. D2I's advantage is specific to reasoning tasks where the structure is an intermediate scaffold. D2I’s success lies in strategically discarding the scaffold to maximize reasoning flexibility, a strategy Seg-Zero cannot adopt.}
    \item \textcolor{black}{VLM-R1 \citep{shen2025vlm} focused on creating a stable and generalizable R1-style VLM by addressing common instabilities in RL training, greatly improved the training stability and scalability of the R1 framework, providing a more robust foundation for any subsequent deliberate training. Despite improving generalization in training, it remains a coupled D2D model. D2I proves that true OOD robustness is achieved not merely by stabilizing the training, but by strategically decoupling the acquired skill from the constraining structure at inference time.}
\end{enumerate}

\section{Effect of Model Scale}  \label{ap:model_size}
%-----------------------------------------------------------------------

\textcolor{black}{This part presents our study on cross-scale generalization, demonstrating how the performance of the D2I framework is affected by variations in model size. Results on Qwen2.5-VL-3B \citep{bai2025qwen2} are shown in Table~\ref{results_3b}. D2I significantly improves the performance of the smaller 3B model, achieving substantial relative gains. This confirms that the method is potent enough to boost even models with limited capacity.}

\begin{table*}[t]

\centering

	\resizebox{\linewidth}{!}{
		\begin{tabular}{lccccccccc}
			\toprule
            % \arrayrulecolor{black}
			 \multirow{2}{*}{\bf Method} & \multicolumn{1}{c}{\textbf{In-domain}} & \multicolumn{3}{c}{\textbf{Out-of-domain (Math)}} & \multicolumn{5}{c}{\textbf{Out-of-domain (General)}} \\ 
               \cmidrule(r){2-2}  \cmidrule(r){3-5} \cmidrule(r){6-10} \noalign{\smallskip} 
             % \hhline{~---------}
             % \noalign{\vskip.2pt}
			  & \multicolumn{1}{c}{\textbf{GEOQA-8K}} & \textbf{\makecell{MathVerse \\ (mini)}} & \textbf{\makecell{MathVista \\ (mini)}} & \textbf{\makecell{MATH-Vision \\ (full)}} & \textbf{\makecell{MME \\ (sum)}} & \textbf{\makecell{MMVet \\ (turbo)}} & \textbf{\makecell{MMMU \\ (val)}} & \textbf{SEED} & \textbf{POPE} \\ \hline

              \rowcolor{blue!5} \multicolumn{10}{c}{\textit{Baselines with the Same Experiment Setting}}\\ \hline
              Qwen2.5-VL-3B$^{\star}$ & \multicolumn{1}{c}{35.3} &38.2&61.9&17.9& 2102.8 &60.2& 52.9 &62.4& 71.0 \\
              \ \ \ \ w/ GRPO & \multicolumn{1}{c}{46.1} &37.1&56.3&17.7& 2154.6 &57.0& 53.1 &60.9& 69.8 \\
              \ \ \ \ w/ GRPO$^{\dagger}$ & \multicolumn{1}{c}{46.4} &37.5&57.2&16.2& 2132.5 &57.2& 52.4 &61.2& 70.3 \\
              \hline
              \rowcolor{blue!5} \multicolumn{10}{c}{\textit{Our Methods}}\\ \hline
              \ \ \ \ w/ D2D$_{loc}$ (ours) & \multicolumn{1}{c}{48.4} &42.4&61.0&\underline{19.4}& 2170.2 &\textbf{61.7}& \underline{54.0} &60.8& 70.0 \\
              \ \ \ \ w/ D2I$_{loc}$ (ours) & \multicolumn{1}{c}{\underline{50.2}} &\underline{43.1}&\textbf{64.2}&\textbf{20.9}& 2146.3 &59.9& 52.6 &61.9& 70.4 \\
              % \rowcolor{gray!20}
              % \ \ \ \ ${\Delta}_{base}$ & \multicolumn{1}{c}{$+$14.0} & $+$2.9 & $+$1.7 & $+$5.0 & $+$4.5 & $+$0.8 & $+$1.8 & $+$1.0 & $+$0.8 \\
              % \rowcolor{gray!20}
              % \ \ \ \ ${\Delta}_{grpo\dagger}$ & \multicolumn{1}{c}{$+$7.5} & $-$0.2 & $+$0.4 & $+$7.5 & $+$99.5 & $+$9.3 & $+$2.0 & $+$0.3 & $+$1.0 \\
              \hline
              \ \ \ \ w/ D2D$_{jus}$ (ours) & \multicolumn{1}{c}{40.3} &40.2&60.3&18.5& 2140.8 &\underline{61.3}& 53.8 &60.2& 69.5 \\
              \ \ \ \ w/ D2I$_{jus}$ (ours) & \multicolumn{1}{c}{\textbf{52.1}} &\textbf{43.7}&\underline{63.2}&17.3& 2129.0 &55.7& 53.1 & 61.0 & 67.6 \\
              % \rowcolor{gray!20}\ \ \ \ ${\Delta}_{base}$ & \multicolumn{1}{c}{$+$18.4} & $+$5.6 & $+$1.5 & $-$1.6 & $+$36.4 & $-$0.2 & $+$2.1 & $+$0.8 & $-$0.2 \\
              % \rowcolor{gray!20}\ \ \ \ ${\Delta}_{grpo\dagger}$ & \multicolumn{1}{c}{$+$11.9} & $+$2.5 & $+$0.2 & $+$0.9 & $+$131.4 & $+$8.3 & $+$2.3 & $+$0.1 & $+$0.0 \\
              \hline
              \ \ \ \ w/ D2D$_{par}$ (ours) & \multicolumn{1}{c}{50.0} &40.3&60.2&16.9& \textbf{2231.5} &60.8& \textbf{56.2} &\underline{64.9}& \textbf{73.5}\\
              \ \ \ \ w/ D2I$_{par}$ (ours) & \multicolumn{1}{c}{\textbf{52.1}} &42.3&62.7&19.1& \underline{2174.1} &57.2& 51.3 &\textbf{66.3}& \underline{72.1} \\
              % \rowcolor{gray!20}\ \ \ \ ${\Delta}_{base}$ & \multicolumn{1}{c}{$+$13.9} & $+$2.4 & $+$4.0 & $+$5.9 & $-$42.8 & $-$1.8 & $+$2.3 & $+$0.5 & $+$0.3 \\
              % \rowcolor{gray!20}\ \ \ \ ${\Delta}_{grpo\dagger}$ & \multicolumn{1}{c}{$+$7.4} & $-$0.7 & $+$2.7 & $+$8.4 & $+$52.2 & $+$6.7 & $+$2.5 & $-$0.2 & $+$0.5 \\
	
            \bottomrule
		\end{tabular}
	}
\caption{\textcolor{black}{Main results of our proposed GEOQA-8K-trained D2I method on both in-domain and out-of-domain test sets. ${\Delta}_{base}$ represents the improvement of D2I over Qwen2.5-VL-3B, while ${\Delta}_{grpo\dagger}$ represents the improvement over Qwen2.5-VL-3B w/ GRPO$^{\dagger}$. $^{\star}$ denotes our re-implementation. All the experiments are conducted on Qwen2.5-VL-3B.}}
\label{results_3b}
\end{table*}

\section{Effect of MLLM Backbone}  \label{ap:backbone}
%-----------------------------------------------------------------------
\textcolor{black}{We also study cross-architecture generalization, demonstrating how the performance of the D2I framework is affected by variations in different MLLM backbone. Results on InternVL2.5-8B \citep{chen2024internvl} are shown in Table~\ref{results_8b}. Applying D2I training to the InternVL2.5-8B architecture, which utilizes a different visual encoder and MLLM family, resulted in similar performance uplifts on key OOD reasoning benchmarks. This strongly validates that D2I is a generalizable training paradigm, not a tuning trick specific to the Qwen family.}

\begin{table*}[t]

\centering

	\resizebox{\linewidth}{!}{
		\begin{tabular}{lccccccccc}
			\toprule
            % \arrayrulecolor{black}
			 \multirow{2}{*}{\bf Method} & \multicolumn{1}{c}{\textbf{In-domain}} & \multicolumn{3}{c}{\textbf{Out-of-domain (Math)}} & \multicolumn{5}{c}{\textbf{Out-of-domain (General)}} \\ 
               \cmidrule(r){2-2}  \cmidrule(r){3-5} \cmidrule(r){6-10} \noalign{\smallskip} 
             % \hhline{~---------}
             % \noalign{\vskip.2pt}
			  & \multicolumn{1}{c}{\textbf{GEOQA-8K}} & \textbf{\makecell{MathVerse \\ (mini)}} & \textbf{\makecell{MathVista \\ (mini)}} & \textbf{\makecell{MATH-Vision \\ (full)}} & \textbf{\makecell{MME \\ (sum)}} & \textbf{\makecell{MMVet \\ (turbo)}} & \textbf{\makecell{MMMU \\ (val)}} & \textbf{SEED} & \textbf{POPE} \\ \hline

              \rowcolor{blue!5} \multicolumn{10}{c}{\textit{Baselines with the Same Experiment Setting}}\\ \hline
              InternVL2.5-8B$^{\star}$ & \multicolumn{1}{c}{40.8} &37.2&61.0&19.2& 2298.1 &61.9& 54.2 &64.2& 84.8 \\
              \ \ \ \ w/ GRPO & \multicolumn{1}{c}{52.7} &41.8&67.1&20.8& 2319.3 &\underline{64.7}& 59.9 &67.1& 88.2 \\
              \ \ \ \ w/ GRPO$^{\dagger}$ & \multicolumn{1}{c}{57.4} &45.6&67.8&19.3& 2322.8 &60.4& 60.1 &66.4& 85.9 \\
              \hline
              \rowcolor{blue!5} \multicolumn{10}{c}{\textit{Our Methods}}\\ \hline
              \ \ \ \ w/ D2D$_{loc}$ (ours) & \multicolumn{1}{c}{54.2} &44.9&69.2&21.0& \textbf{2358.2} &62.8& \underline{63.7} &\underline{69.1}& \underline{89.2} \\
              \ \ \ \ w/ D2I$_{loc}$ (ours) & \multicolumn{1}{c}{\textbf{59.2}} &\textbf{47.1}&\textbf{69.8}&\underline{22.4}& 2273.6 &63.3& \textbf{64.0} &68.7& 88.5 \\
              % \rowcolor{gray!20}
              % \ \ \ \ ${\Delta}_{base}$ & \multicolumn{1}{c}{$+$14.0} & $+$2.9 & $+$1.7 & $+$5.0 & $+$4.5 & $+$0.8 & $+$1.8 & $+$1.0 & $+$0.8 \\
              % \rowcolor{gray!20}
              % \ \ \ \ ${\Delta}_{grpo\dagger}$ & \multicolumn{1}{c}{$+$7.5} & $-$0.2 & $+$0.4 & $+$7.5 & $+$99.5 & $+$9.3 & $+$2.0 & $+$0.3 & $+$1.0 \\
              \hline
              \ \ \ \ w/ D2D$_{jus}$ (ours) & \multicolumn{1}{c}{53.8} &44.2&67.5&17.9& 2301.6 &\textbf{65.3}& 62.3 &\textbf{69.8}& \textbf{89.3} \\
              \ \ \ \ w/ D2I$_{jus}$ (ours) & \multicolumn{1}{c}{50.3} &45.2&64.7&21.4& 2326.7 &61.0& 61.8 & 66.2 & 82.0 \\
              % \rowcolor{gray!20}\ \ \ \ ${\Delta}_{base}$ & \multicolumn{1}{c}{$+$18.4} & $+$5.6 & $+$1.5 & $-$1.6 & $+$36.4 & $-$0.2 & $+$2.1 & $+$0.8 & $-$0.2 \\
              % \rowcolor{gray!20}\ \ \ \ ${\Delta}_{grpo\dagger}$ & \multicolumn{1}{c}{$+$11.9} & $+$2.5 & $+$0.2 & $+$0.9 & $+$131.4 & $+$8.3 & $+$2.3 & $+$0.1 & $+$0.0 \\
              \hline
              \ \ \ \ w/ D2D$_{par}$ (ours) & \multicolumn{1}{c}{54.0} &42.9&68.1&20.4& \underline{2340.9} &62.7& 60.9 &67.5& 88.7\\
              \ \ \ \ w/ D2I$_{par}$ (ours) & \multicolumn{1}{c}{\underline{57.7}} &\underline{45.8}&\underline{69.3}&\textbf{23.9}& 2306.3 &60.4& 60.4 &67.9& 89.0 \\
              % \rowcolor{gray!20}\ \ \ \ ${\Delta}_{base}$ & \multicolumn{1}{c}{$+$13.9} & $+$2.4 & $+$4.0 & $+$5.9 & $-$42.8 & $-$1.8 & $+$2.3 & $+$0.5 & $+$0.3 \\
              % \rowcolor{gray!20}\ \ \ \ ${\Delta}_{grpo\dagger}$ & \multicolumn{1}{c}{$+$7.4} & $-$0.7 & $+$2.7 & $+$8.4 & $+$52.2 & $+$6.7 & $+$2.5 & $-$0.2 & $+$0.5 \\
	
            \bottomrule
		\end{tabular}
	}
\caption{\textcolor{black}{Main results of our proposed GEOQA-8K-trained D2I method on both in-domain and out-of-domain test sets. ${\Delta}_{base}$ represents the improvement of D2I over InternVL2.5-8B, while ${\Delta}_{grpo\dagger}$ represents the improvement over InternVL2.5-8B w/ GRPO$^{\dagger}$. $^{\star}$ denotes our re-implementation. All the experiments are conducted on InternVL2.5-8B.}}
\label{results_8b}
\end{table*}

\section{Training of SFT-Only and SFT-RL Models} 

\label{ap:sft}

Specifically, we construct two distinct SFT datasets to train SFT-only models as shown in Table~\ref{ablation}. The \textit{loc-sft} dataset focuses on predicting bounding box coordinates for crucial regions from mathematical images, enabling the model to learn accurate spatial localization. The SFT model trained on this dataset is referred to as SFT$_{loc}$. The \textit{par-sft} dataset is designed for mathematical image parsing, aimed at teaching the model to generate correct parsing outputs; we denote the resulting model as SFT$_{par}$. Building upon these SFT models, we further update them using our proposed RL framework to form the SFT-RL models. By applying our method to SFT$_{loc}$, we obtain two models under corresponding deliberate reasoning strategy: SFT-D2D$_{loc}$ and SFT-D2I$_{loc}$. A similar procedure is applied to SFT$_{par}$, yielding SFT-D2D$_{par}$ and SFT-D2I$_{par}$.

\section{Entropy Distribution}  \label{ap:entropy}

\begin{figure*}[t]
	\centering
        \includegraphics[width=\textwidth]{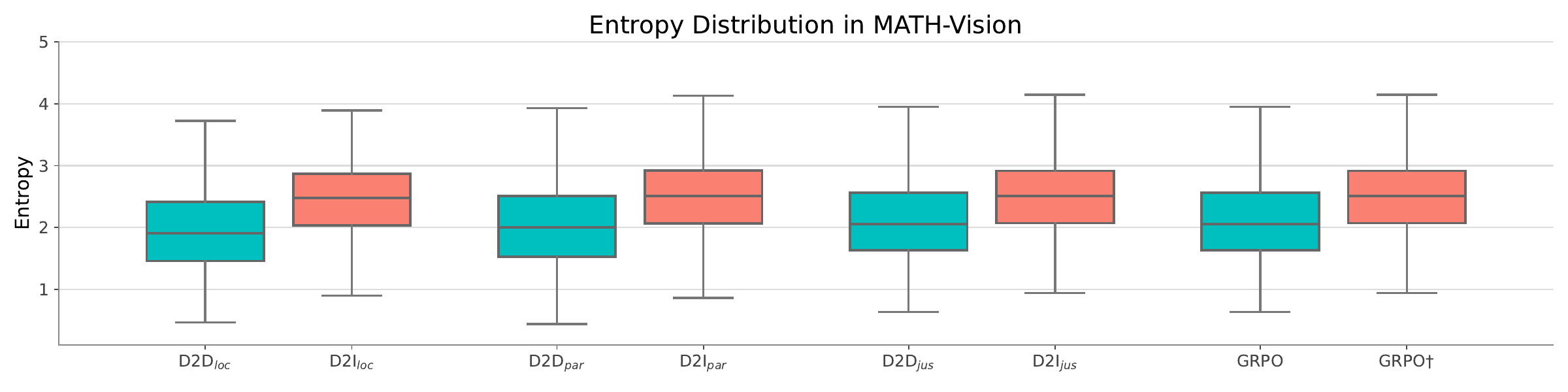}
	\caption{Entropy distribution results on MATH-Vision dataset.}
	\label{fig_entropy_vision}
\end{figure*}

\begin{figure*}[t]
	\centering
        \includegraphics[width=\textwidth]{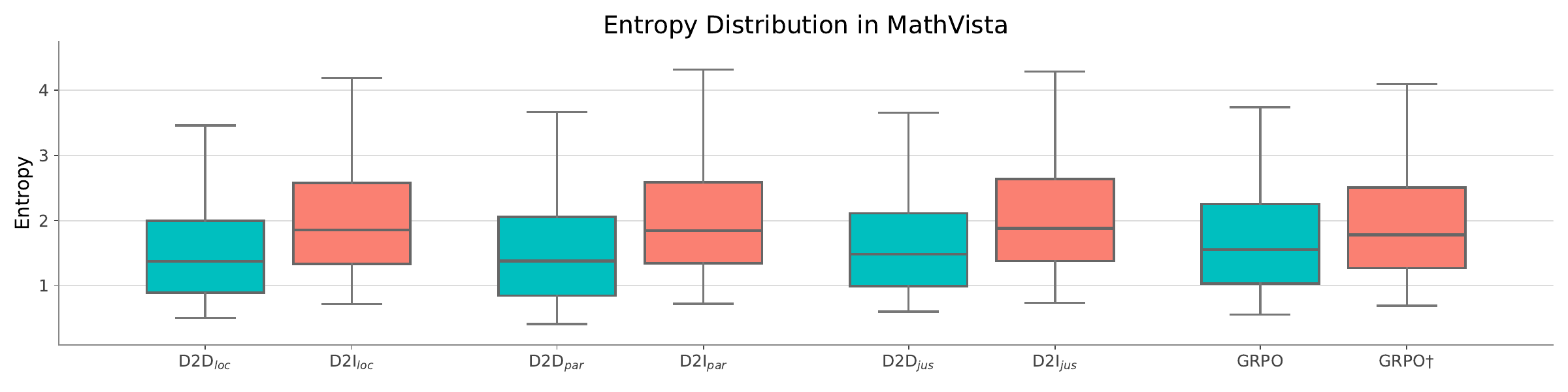}
	\caption{Entropy distribution results on MathVista dataset.}
	\label{fig_entropy_vista}
\end{figure*}

We show the entropy distribution on MATH-Vision in Fig.~\ref{fig_entropy_vision} and on MathVista in Fig.~\ref{fig_entropy_vista}.

\section{Token Distribution Shift}  \label{ap:tokenshift}

\begin{figure*}[t]
	\centering
        \includegraphics[width=\textwidth]{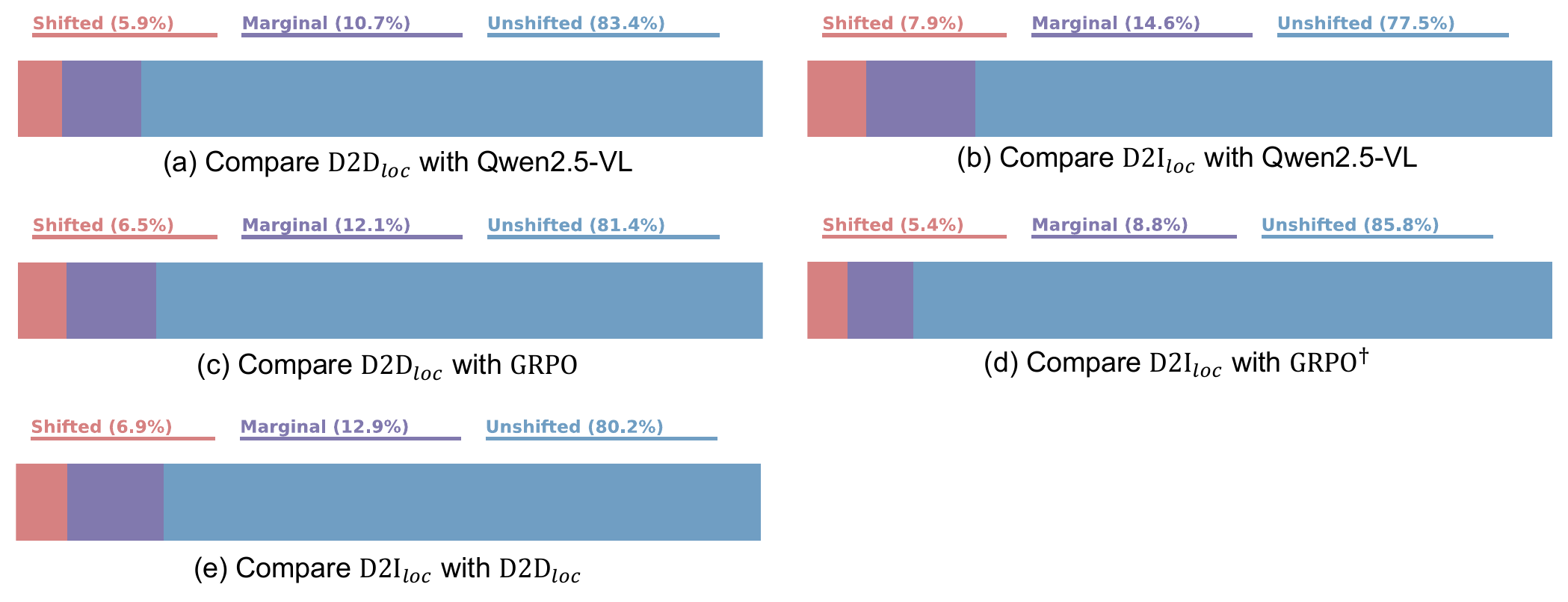}
	\caption{Token Distribution Shift of LOC on MathVerse dataset.}
	\label{shift_box_verse}
\end{figure*}

\begin{figure*}[t]
	\centering
        \includegraphics[width=\textwidth]{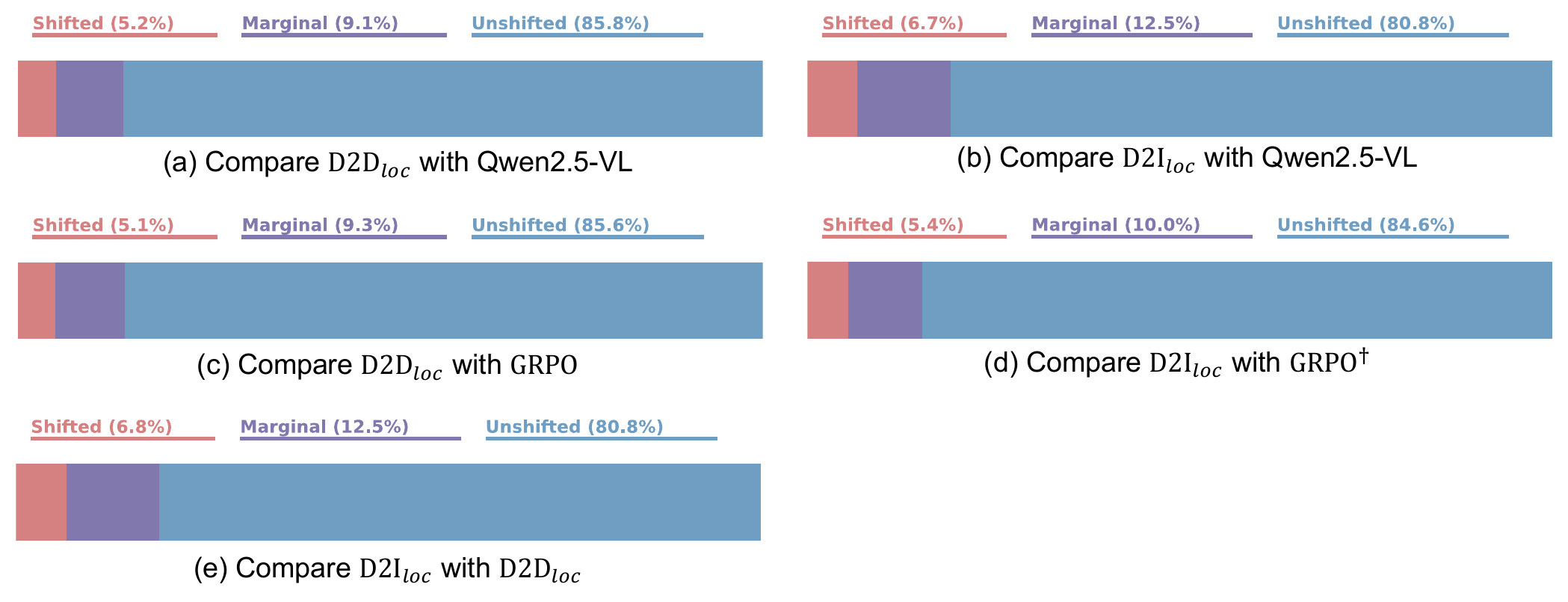}
	\caption{Token Distribution Shift of LOC on MathVista dataset.}
	\label{shift_box_vista}
\end{figure*}

\begin{figure*}[t]
	\centering
        \includegraphics[width=\textwidth]{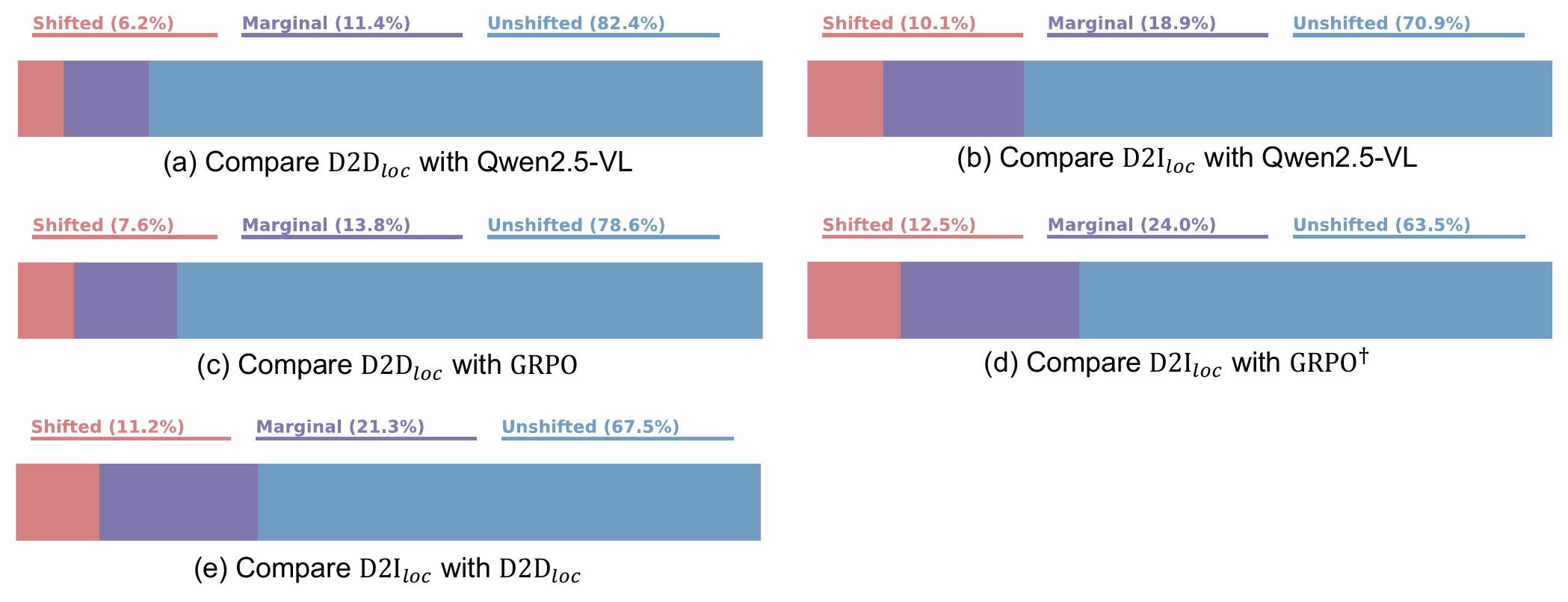}
	\caption{Token Distribution Shift of LOC on MATH-Vision dataset.}
	\label{shift_box_vision}
\end{figure*}

\begin{figure*}[t]
	\centering
        \includegraphics[width=\textwidth]{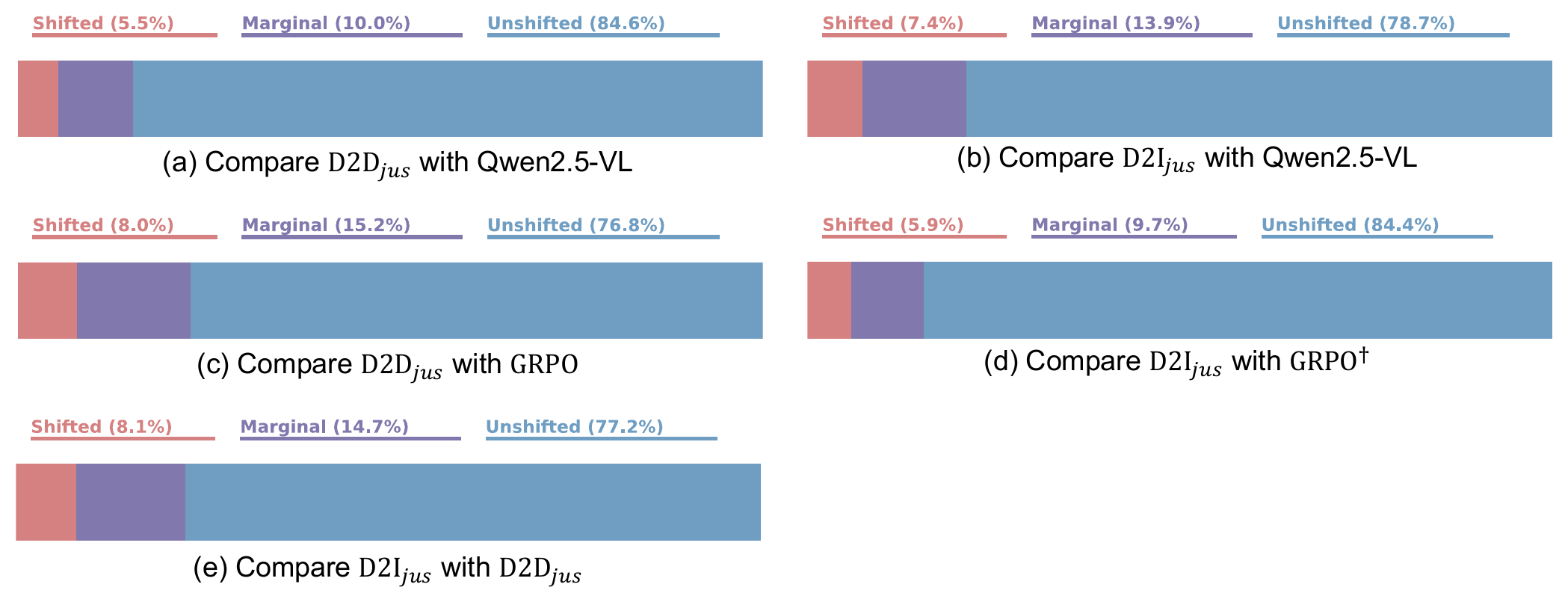}
	\caption{Token Distribution Shift of JUS on MathVerse dataset.}
	\label{shift_ctext_verse}
\end{figure*}

\begin{figure*}[t]
	\centering
        \includegraphics[width=\textwidth]{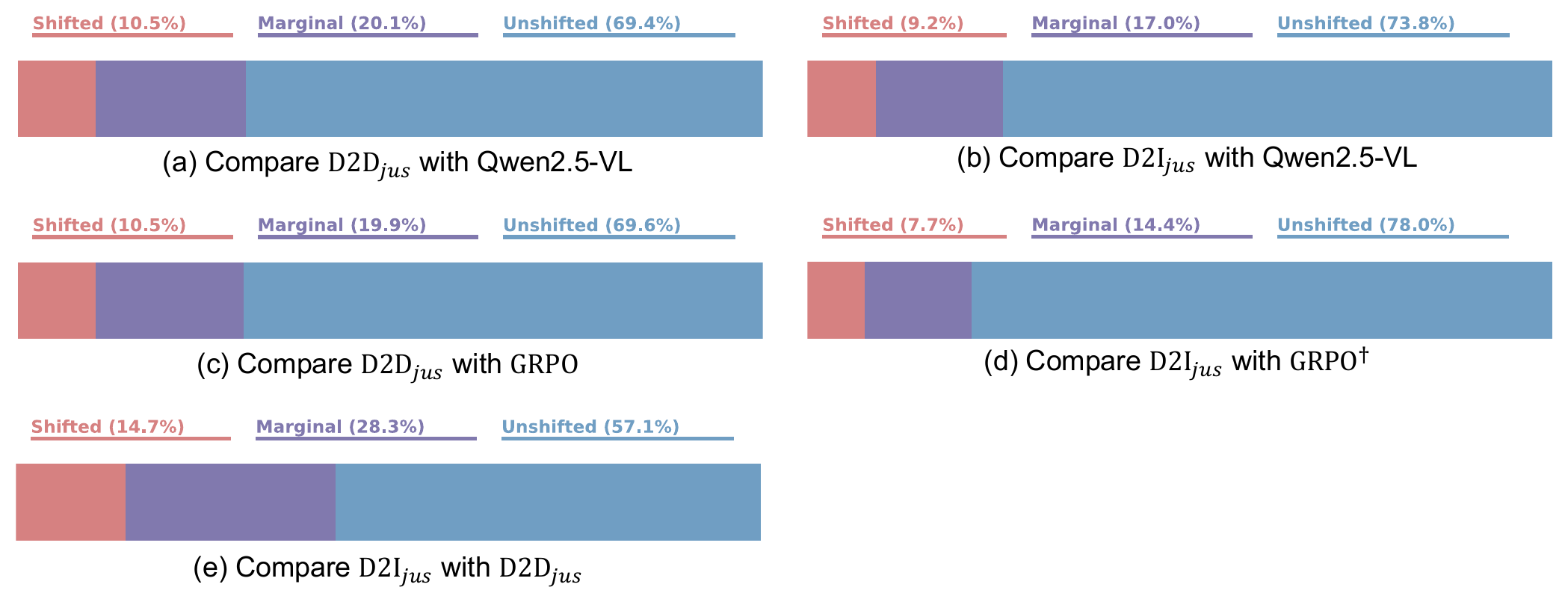}
	\caption{Token Distribution Shift of JUS on MathVista dataset.}
	\label{shift_ctext_vista}
\end{figure*}

\begin{figure*}[t]
	\centering
        \includegraphics[width=\textwidth]{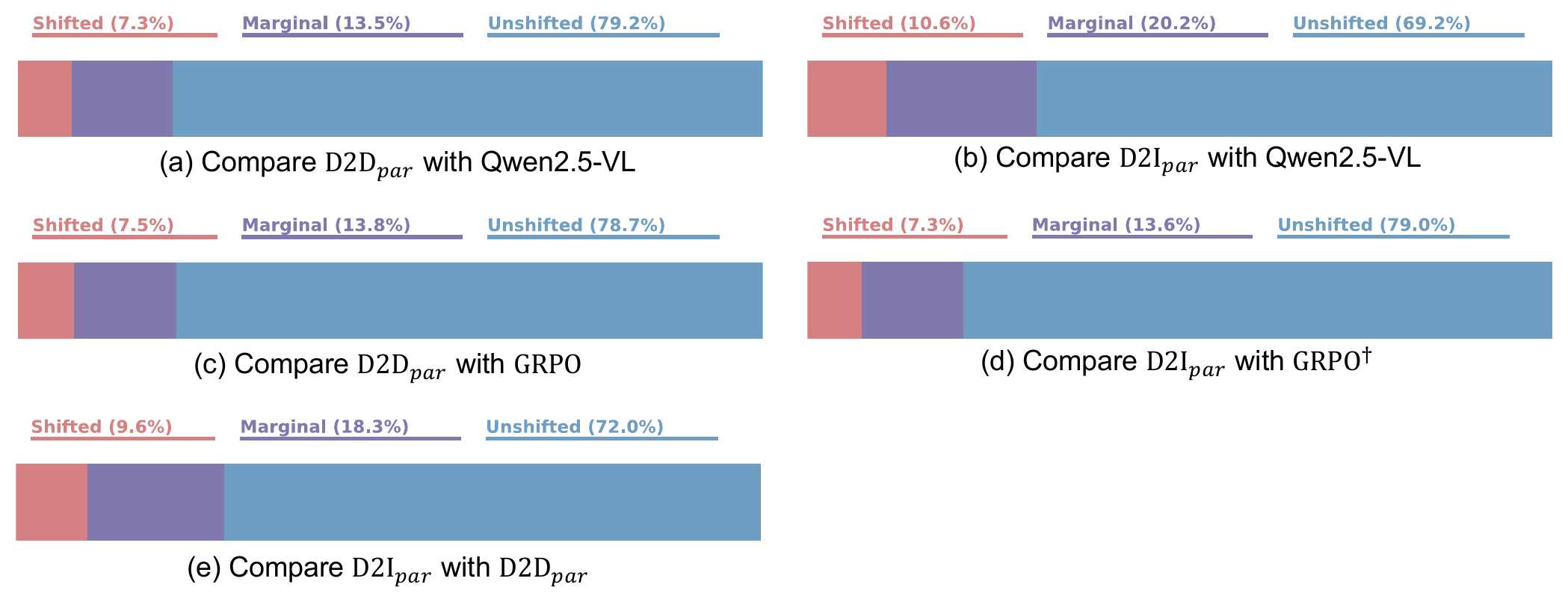}
	\caption{Token Distribution Shift of PAR on MathVerse dataset.}
	\label{shift_par_verse}
\end{figure*}

\begin{figure*}[t]
	\centering
        \includegraphics[width=\textwidth]{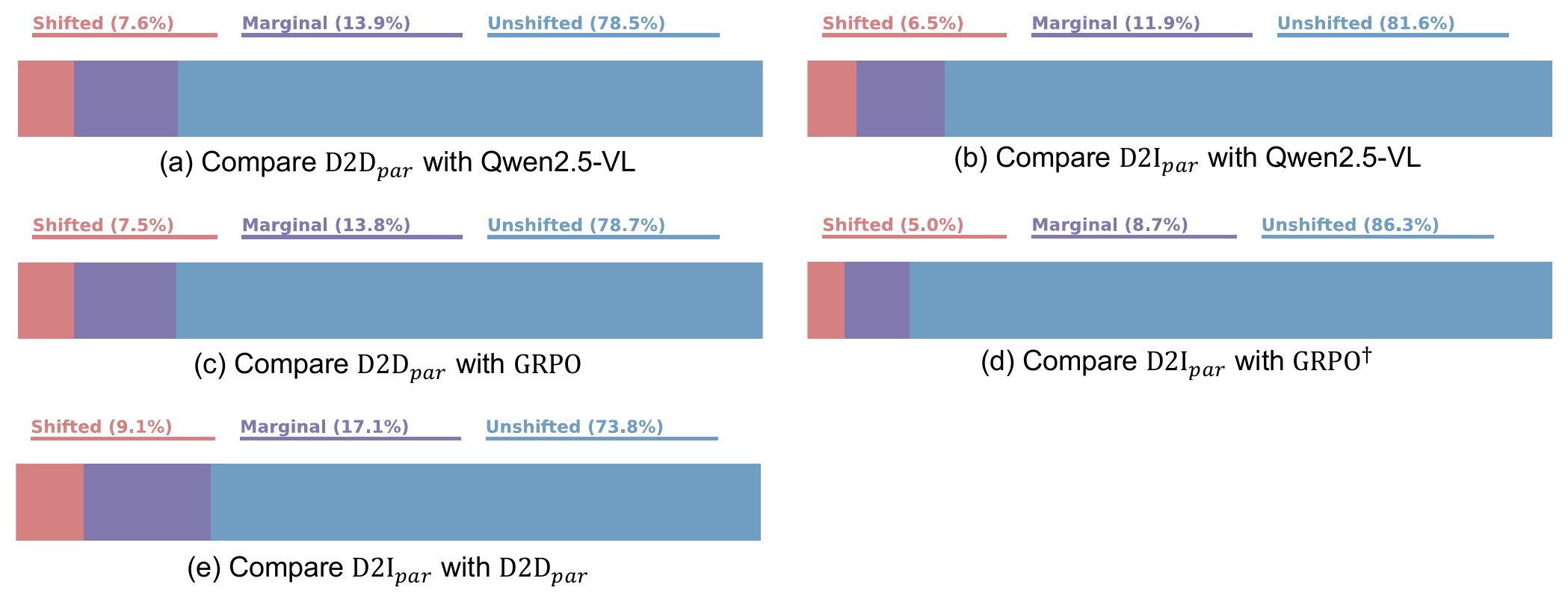}
	\caption{Token Distribution Shift of PAR on MathVista dataset.}
	\label{shift_par_vista}
\end{figure*}

\begin{figure*}[t]
	\centering
        \includegraphics[width=\textwidth]{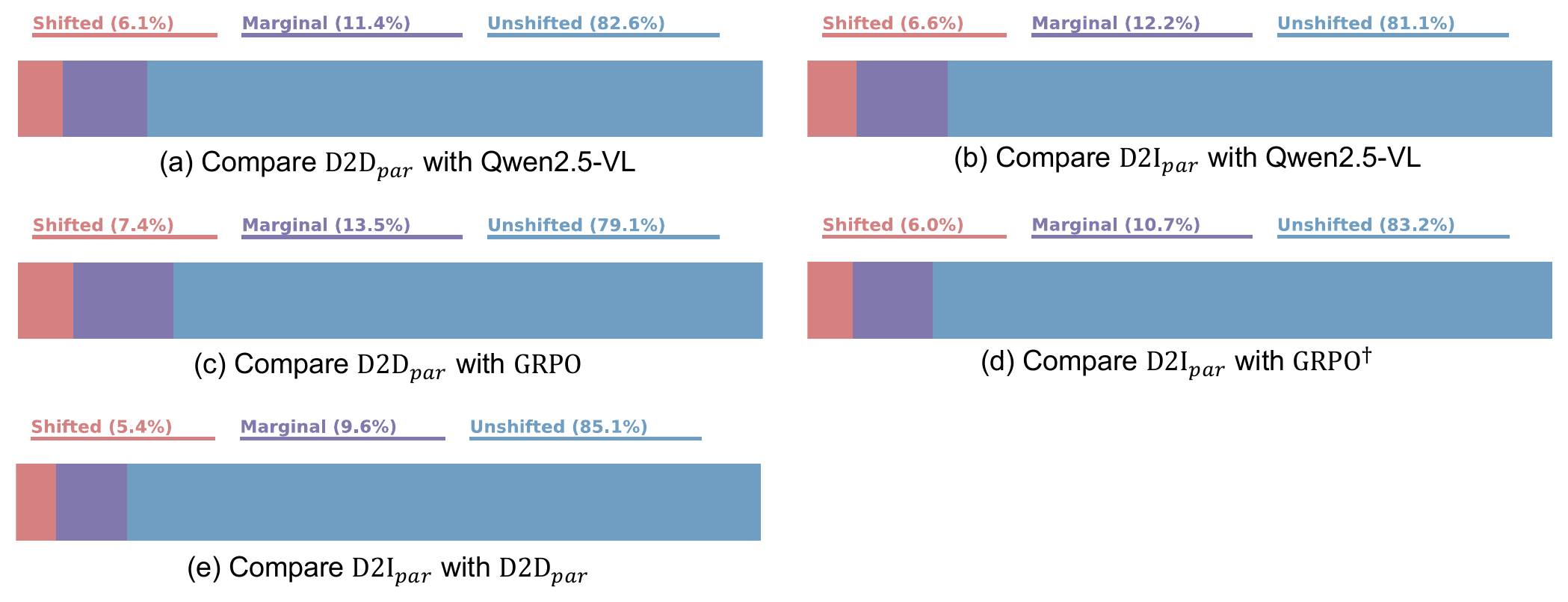}
	\caption{Token Distribution Shift of PAR on MATH-Vision dataset.}
	\label{shift_par_vision}
\end{figure*}

\begin{figure*}[t]
	\centering
        \includegraphics[width=\textwidth]{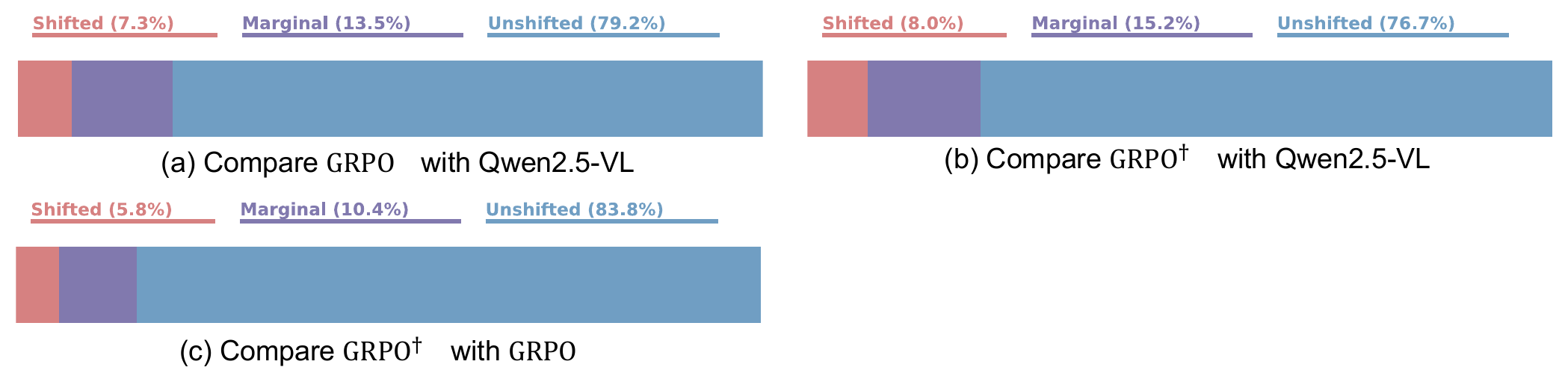}
	\caption{Token Distribution Shift of GRPO on MathVerse dataset.}
	\label{shift_grpo_verse}
\end{figure*}

\begin{figure*}[t]
	\centering
        \includegraphics[width=\textwidth]{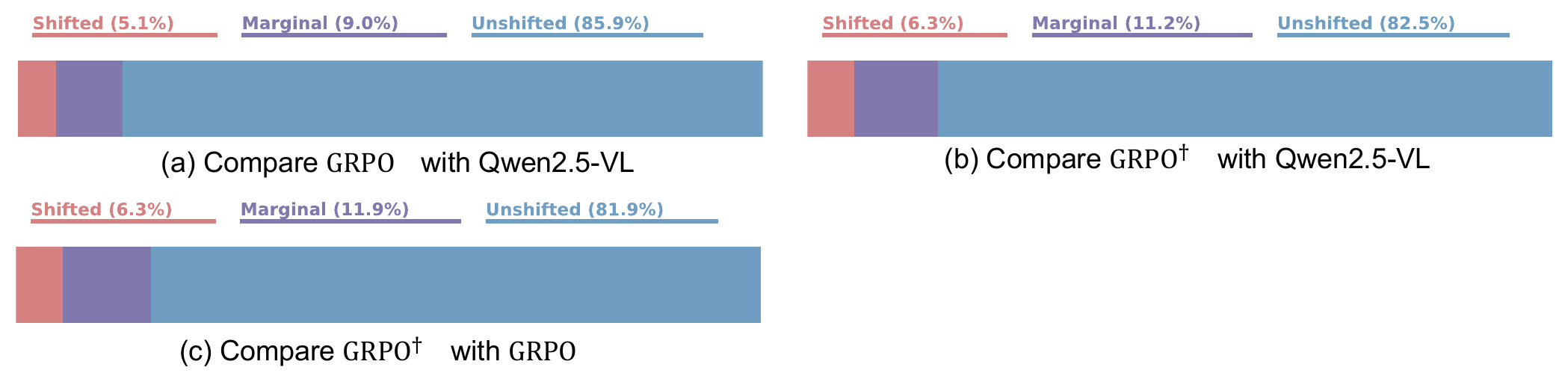}
	\caption{Token Distribution Shift of GRPO on MathVista dataset.}
	\label{shift_grpo_vista}
\end{figure*}

\begin{figure*}[t]
	\centering
        \includegraphics[width=\textwidth]{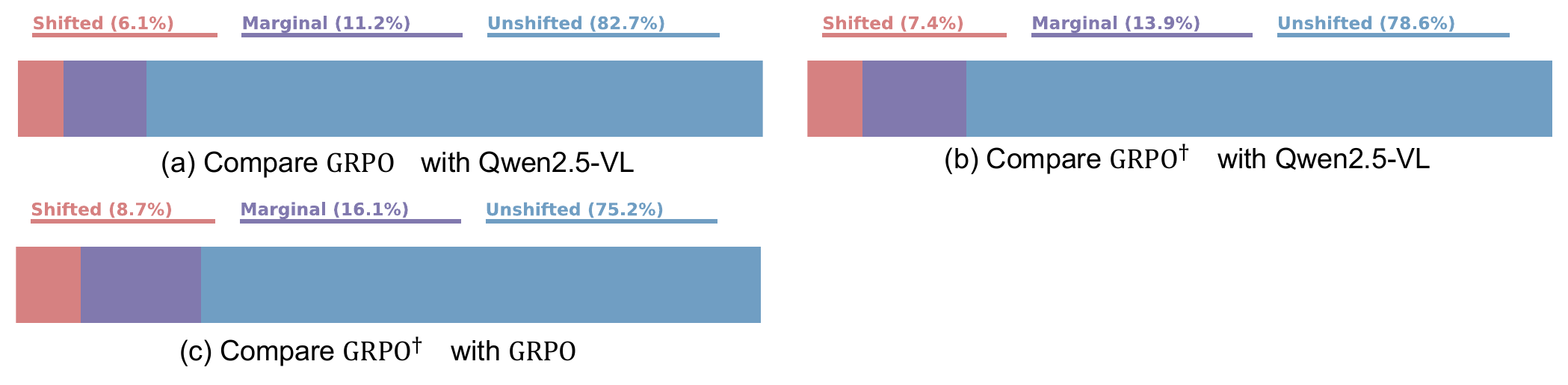}
	\caption{Token Distribution Shift of GRPO on MATH-Vision dataset.}
	\label{shift_grpo_vision}
\end{figure*}

We show the token distribution shift results in Fig.s~\ref{shift_box_verse}, \ref{shift_box_vista}, \ref{shift_box_vision}, \ref{shift_ctext_verse}, \ref{shift_ctext_vista}, \ref{shift_par_verse}, \ref{shift_par_vista}, \ref{shift_par_vision}, \ref{shift_grpo_verse}, \ref{shift_grpo_vista}, \ref{shift_grpo_vision}.

\section{Word Cloud Visualizations }  \label{ap:cloud}

\begin{figure*}[t]
	\centering
        \includegraphics[width=\textwidth]{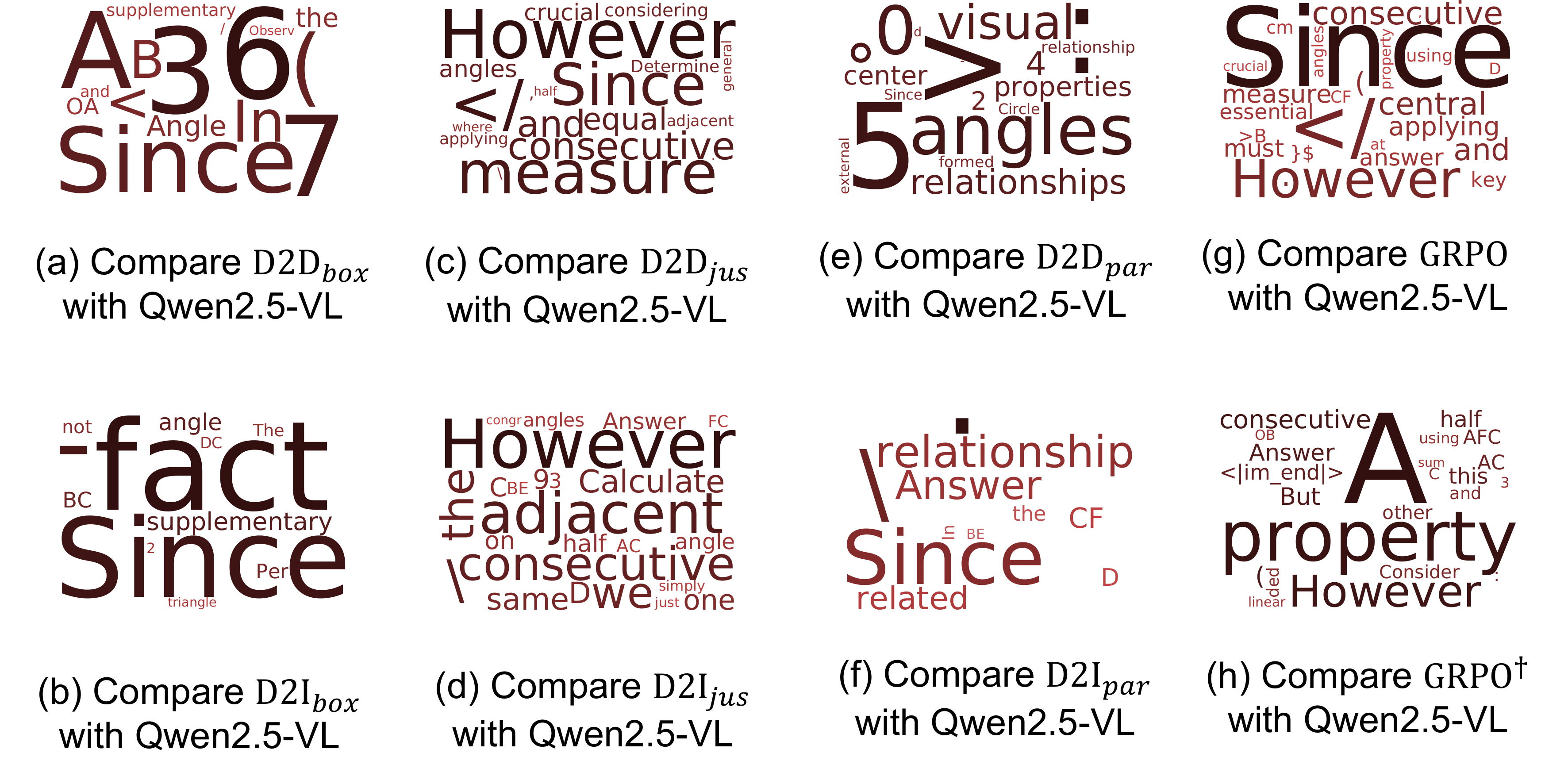}
	\caption{Word cloud visualizations on MathVerse.}
	\label{word1}
\end{figure*}

\begin{figure*}[t]
	\centering
        \includegraphics[width=0.99\textwidth]{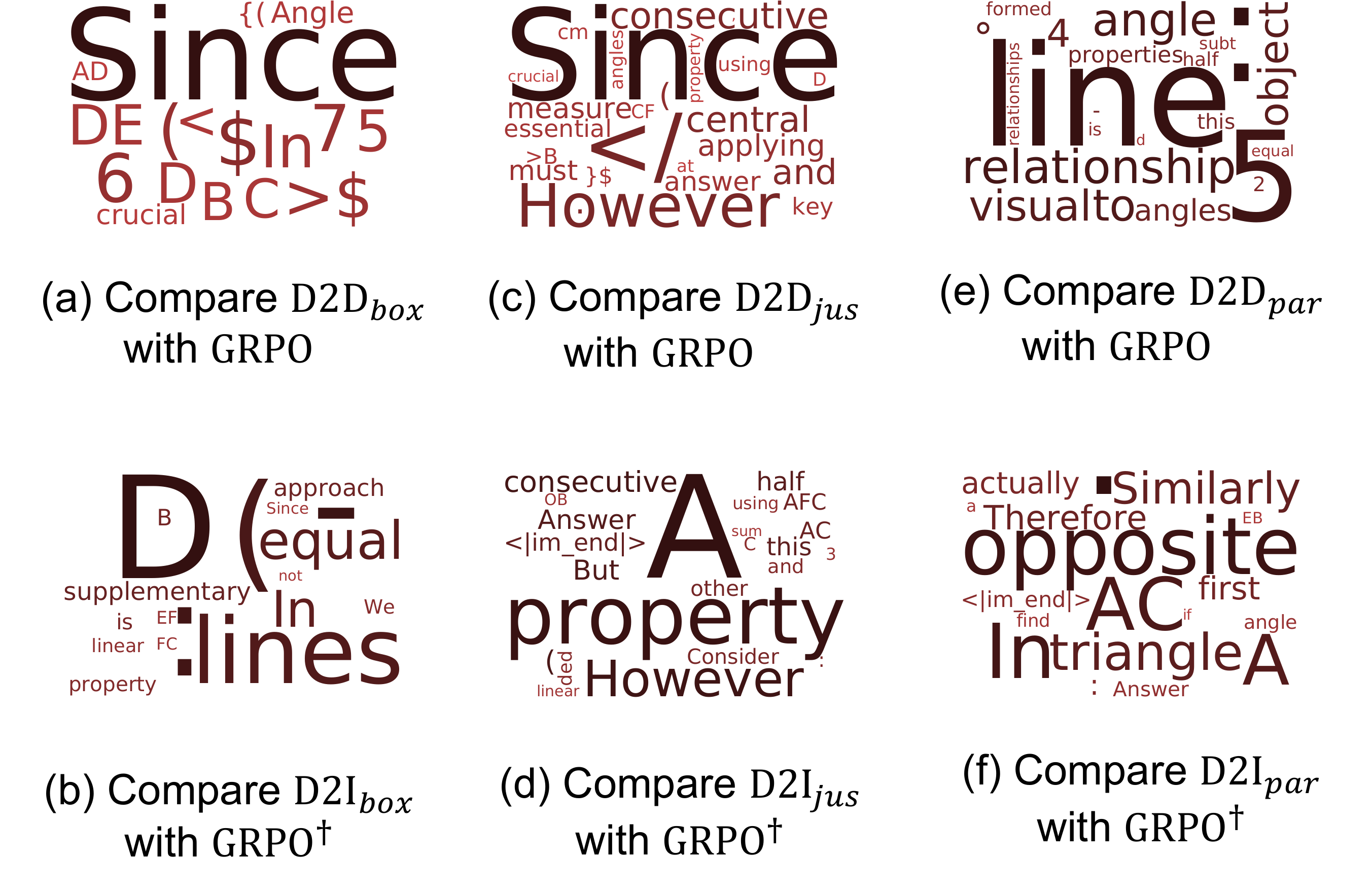}
	\caption{Word cloud visualizations on MathVerse.}
	\label{word2}
\end{figure*}

\begin{figure*}[t]
	\centering
        \includegraphics[width=\textwidth]{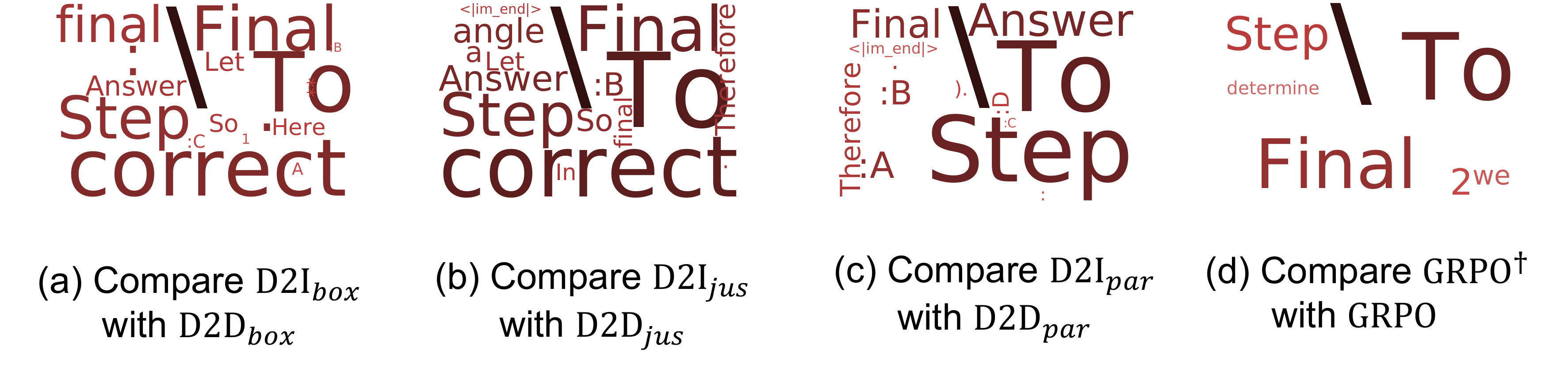}
	\caption{Word cloud visualizations on MathVerse.}
	\label{word3}
\end{figure*}

We also present word cloud visualizations to more intuitively illustrate which specific tokens were altered in MathVerse dataset. Results are shown in Fig.s~\ref{word1},~\ref{word2} and~\ref{word3}.

\section{More Case Analysis}  \label{ap:more_case}
%-----------------------------------------------------------------------

\begin{figure*}[t]
	\centering
    \includegraphics[width=0.95\linewidth]{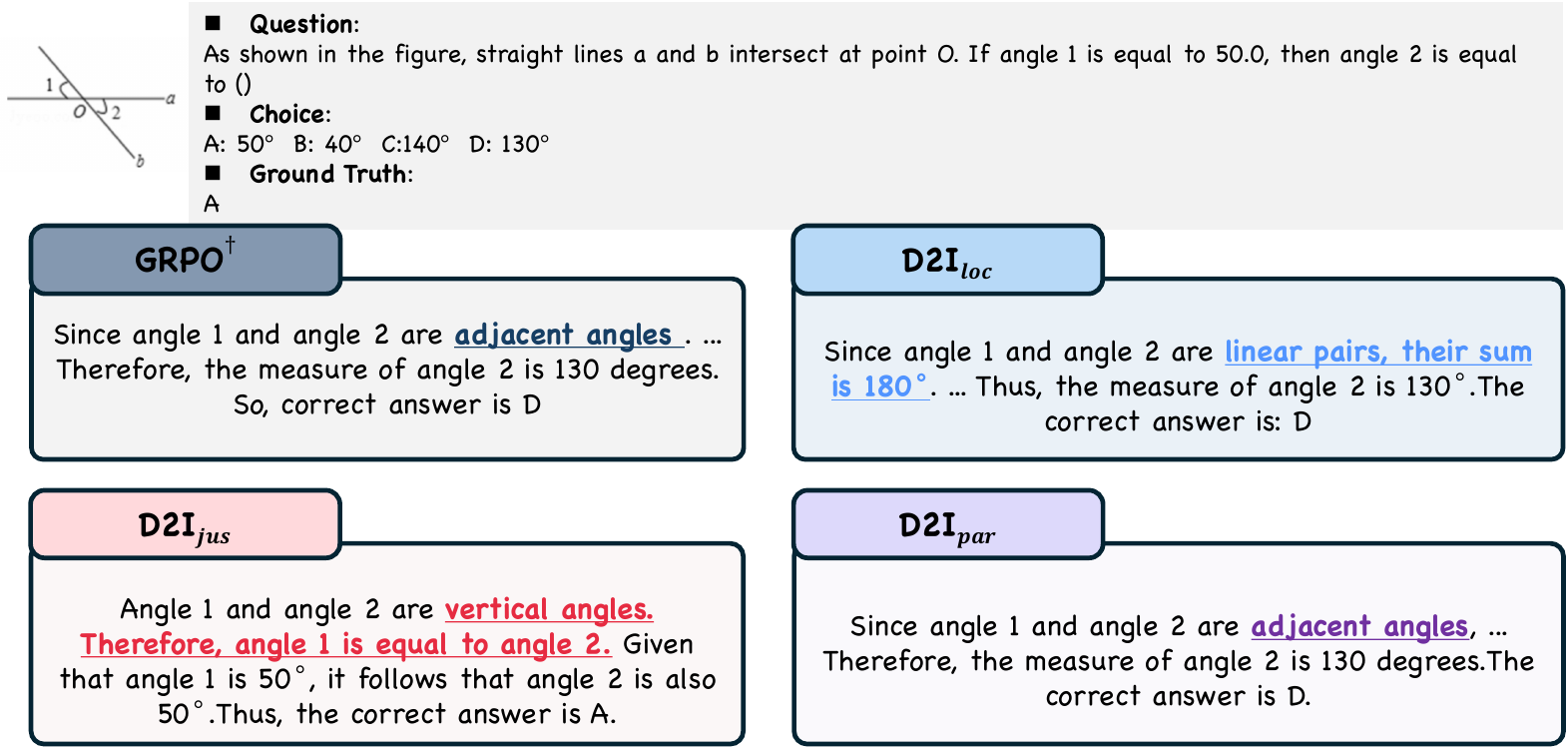}
	\caption{Inference responses to the same math question from models trained with different deliberate reasoning strategies.}
	\label{fig:curve}
\end{figure*}

%-----------------------------------------------------------------------
We analyze a representative failure case where GRPO, D2I$_{loc}$, and D2I$_{par}$ all produce incorrect answers, while D2I$_{jus}$ succeeds. As shown in Fig. \ref{fig:curve}, notably, all four responses share a highly similar structure and format, demonstrating the same reasoning steps. 
However, only the D2I$_{jus}$ model correctly identifies a critical concept in its reasoning trace: the mention of \textit{vertical angles}. This small but crucial difference allows D2I$_{jus}$ to reach the correct final answer, whereas the other models fail. 
This highlights an important observation: \textbf{Once the key visual perceptions are correctly understood, the model can form a complete and accurate reasoning chain, with no explicit output format or structured intermediate representation required}. 
It underscores that understanding and articulating the right concept, rather than strictly enforcing output formats, is often the deciding factor in successful reasoning. 
This case also illustrates the unique strength of D2I$_{jus}$ in aligning with the LLM's natural generation behavior. 

\section{The Use of Large Language Models}

In the preparation of this manuscript, LLMs were used solely for the purpose of text polishing, including grammar correction and stylistic refinement. No content was generated or rewritten with the intention of altering the scientific meaning, originality, or conclusions of the work. All ideas, analyses, and results presented in this paper are entirely the authors' own.

\end{document}